%% file: paper.tex
\documentclass{article}

\newif\ifarxiv
\arxivtrue

\ifarxiv
\newcommand{\thesupplement}{the appendix}
\newcommand{\suppmat}{appendix}
\newcommand{\thesuppmat}{the appendix}
\else
\newcommand{\thesupplement}{the supplement}
\newcommand{\suppmat}{supplemental material}
\newcommand{\thesuppmat}{the supplemental material}
\fi

\usepackage{microtype}
\usepackage{graphicx}
\usepackage{subfigure}
\usepackage{booktabs} %

\usepackage{hyperref}

\usepackage{color}
\usepackage[export]{adjustbox}
\usepackage{float}
\usepackage{url}
\usepackage{xfrac}
\usepackage{xspace}
\usepackage[normalem]{ulem}  %
\usepackage{enumitem}  		%
\usepackage{tikz}			%
\usepackage{calc}			%
\usepackage{collcell}
\usepackage{amsmath,amssymb}
\usepackage{multirow}
\usepackage{titlesec}
\usepackage[utf8]{inputenc}

\usepackage[accepted]{icml2018b}

\newcommand{\ppm}{${}\pm{}$}
\newcommand{\expectation}[1]{\mathbb{E}\{{#1}\}}
\newcommand{\expect}[2]{\mathbb{E}_{#1}\{{#2}\}}
\newcommand{\inputx}{\hat{x}}
\newcommand{\inputy}{\hat{y}}
\DeclareMathOperator*{\argmin}{arg min}
\newcommand{\diff}{\text{d}}

\def\clap#1{\hbox to 0pt{\hss #1\hss}}%
\definecolor{internationalkleinblue}{rgb}{0.0, 0.18, 0.65}	%
\definecolor{olive}{rgb}{0.5, 0.5, 0.0}
\definecolor{maroon}{rgb}{0.69, 0.19, 0.38}
\definecolor{celestialblue}{rgb}{0.29, 0.59, 0.82}
\definecolor{darkgreen}{rgb}{0.0, 0.5, 0.0}

\newcommand{\ImageNet}{\textsc{ImageNet}}
\newcommand{\Kodak}{\textsc{Kodak}}
\newcommand{\BSD}{\textsc{BSD300}}
\newcommand{\SET}{\textsc{Set14}}
\newcommand{\papertitle}{Noise2Noise: Learning Image Restoration without Clean Data}

\input{figures}

\icmltitlerunning{\papertitle}

\begin{document}

\twocolumn[
\icmltitle{\papertitle}

\icmlsetsymbol{equal}{*}

\begin{icmlauthorlist}
\icmlauthor{Jaakko Lehtinen}{nv,aa}
\icmlauthor{Jacob Munkberg}{nv}
\icmlauthor{Jon Hasselgren}{nv}
\icmlauthor{Samuli Laine}{nv}
\icmlauthor{Tero Karras}{nv}
\icmlauthor{Miika Aittala}{mit}
\icmlauthor{Timo Aila}{nv}
\end{icmlauthorlist}
\icmlaffiliation{nv}{NVIDIA}
\icmlaffiliation{mit}{MIT CSAIL}
\icmlaffiliation{aa}{Aalto University}

\icmlcorrespondingauthor{Jaakko Lehtinen}{jlehtinen@nvidia.com}

\icmlkeywords{Machine Learning, ICML}

\vskip 0.3in
]

\printAffiliationsAndNotice{}  %

\begin{abstract}
We apply basic statistical reasoning to signal reconstruction by machine learning -- learning to map corrupted observations to clean signals -- with a simple and powerful conclusion: it is possible to learn to restore images by only looking at corrupted examples, at performance at and sometimes exceeding training using clean data, without explicit image priors or likelihood models of the corruption. In practice, we show that a single model learns photographic noise removal, denoising synthetic Monte Carlo images, and reconstruction of undersampled MRI scans -- all corrupted by different processes -- based on noisy data only.
\end{abstract}

\section{Introduction}
\label{sec:intro}
\input{intro.tex}

\section{Practical Experiments}
\label{sec:applications}

We now experimentally study the practical properties of noisy-target training. %
We start with simple noise distributions (Gaussian, Poisson, Bernoulli) in Sections~\ref{sec:gaussian} and~\ref{sec:synthetic}, and continue to the much harder, analytically intractable Monte Carlo image synthesis noise (Section~\ref{sec:mc}). In Section~\ref{sec:mri}, we show that image reconstruction from sub-Nyquist spectral samplings in magnetic resonance imaging (MRI) can be learned from corrupted observations only. %

\subsection{Additive Gaussian Noise}
\label{sec:gaussian}

\figconvergence

We will first study the effect of corrupted targets using synthetic additive Gaussian noise. %
As the noise has zero mean, we use the $L_2$ loss for training to recover the mean.

Our baseline is a recent state-of-the-art method "RED30" \cite{Mao2016b}, a 30-layer hierarchical residual network with 128 feature maps, which has been demonstrated to be very effective in a wide range of image restoration tasks, including Gaussian noise.
We train the network using 256$\times$256-pixel crops drawn from the 50k images in the \ImageNet~validation set.
We furthermore randomize the noise standard deviation $\sigma \in [0,50]$ separately for each training example, i.e., the network has to estimate the magnitude of noise while removing it (``blind'' denoising). 

\tabpsnr

We use three well-known datasets: \BSD~\cite{BSD300}, \SET~\cite{Zeyde2010}, and \Kodak\footnote{http://r0k.us/graphics/kodak/}. As summarized in Table~\ref{tab:psnr}, the behavior is qualitatively similar in all three sets, and thus we discuss the averages. 
When trained using the standard way with clean targets (Equation~\ref{eq:regression}), RED30 achieves  31.63\ppm0.02\,dB with $\sigma=25$. The confidence interval was computed by sampling five random initializations. 
The widely used benchmark denoiser BM3D \cite{Dabov2007} gives $\sim$0.7\,dB worse results.
When we modify the training to use noisy targets (Equation~\ref{eq:n2n}) instead, the denoising performance remains equally good. Furthermore, the training converges just as quickly, as shown in Figure~\ref{fig:convergence}a. This leads us to conclude that clean targets are unnecessary in this application. This perhaps surprising observation holds also with different networks and network capacities. 
Figure~\ref{fig:simple}a shows an example result. 

For all further tests, we switch from RED30 to a shallower U-Net \cite{Ronneberger2015} that is roughly 10$\times$ faster to train and gives similar results ($-$0.2\,dB in Gaussian noise). The architecture and training parameters are described in \thesuppmat.

\textbf{Convergence speed}
Clearly, every training example asks for the impossible: there is no way the network could succeed in transforming one instance of the noise to another. Consequently, the training loss does actually not decrease during training, and the loss gradients continue to be quite large. Why do the larger, noisier gradients not affect convergence speed? While the activation gradients are indeed noisy, the weight gradients are in fact relatively clean because Gaussian noise is independent and identically distributed (i.i.d.) in all pixels, and the weight gradients get averaged over $2^{16}$ pixels in our fully convolutional network.

Figure~\ref{fig:convergence}b makes the situation harder by introducing inter-pixel correlation to the noise. This brown additive noise is obtained by blurring white Gaussian noise by a spatial Gaussian filter of different bandwidths and scaling to retain $\sigma=25$.
An example is shown in Figure~\ref{fig:convergence}b. As the correlation increases, the effective averaging of weight gradients decreases, and the weight updates become noisier. This makes the convergence slower, but even with extreme blur, the eventual quality is similar (within 0.1\,dB). %

\textbf{Finite data and capture budget}
The previous studies relied on the availability of infinitely many noisy examples produced by adding synthetic noise to clean images. We now study corrupted vs. clean training data in the realistic scenario of finite data and a fixed capture budget.
Our experiment setup is as follows. Let one ImageNet image with white additive Gaussian noise at $\sigma=25$ correspond to one ``capture unit'' (CU). Suppose that 19 CUs are enough for a clean capture, so that one noisy realization plus the clean version (the average of 19 noisy realizations) consumes 20 CU. Let us fix a total capture budget of, say, 2000 CUs. This budget can be allocated between clean latents ($N$) and noise realizations per clean latent ($M$) such that $N*M=2000$. In the traditional scenario, we have only 100 training pairs ($N=100$, $M=20$): a single noisy realization and the corresponding clean image (= average of 19 noisy images; Figure~\ref{fig:convergence}c, Case 1). We first observe that using the \emph{same} captured data as $100*20*19=38000$ training pairs with corrupted targets --- i.e., for each latent, forming all the $19*20$ possible noisy/clean pairs --- yields notably better results (several .1s of dB) than the traditional, fixed noisy+clean pairs, even if we still only have $N=100$ latents (Figure~\ref{fig:convergence}c, Case 2). Second, we observe that setting $N=1000$ and $M=2$, i.e., increasing the number of clean latents but only obtaining two noisy realizations of each (resulting in 2000 training pairs) yields even better results (again, by several .1s of dB, Figure~\ref{fig:convergence}c, Case 3).

We conclude that for additive Gaussian noise, \emph{corrupted targets offer benefits --- not just the same performance but better --- over clean targets} on two levels: both 1) seeing more realizations of the corruption for the same latent clean image, and 2) seeing more latent clean images, even if just two corrupted realizations of each, are beneficial.

\subsection{Other Synthetic Noises}
\label{sec:synthetic}

We will now experiment with other types of synthetic noise. The training setup is the same as described above.

\textbf{Poisson noise} is the dominant source of noise in photographs. %
While zero-mean, it is harder to remove because it is signal-dependent.
We use the $L_2$ loss, and vary the noise magnitude $\lambda \in [0,50]$ during training. Training with clean targets results in 30.59\ppm0.02\,dB, while noisy targets give an equally good 30.57\ppm0.02\,dB, again at similar convergence speed. A comparison method \cite{Makitalo2011} that first transforms the input Poisson noise into Gaussian (Anscombe transform), then denoises by BM3D, and finally inverts the transform, yields 2\,dB less.

Other effects, e.g., dark current and quantization, are dominated by Poisson noise, can be made zero-mean \cite{Hasinoff2016}, and hence pose no problems for training with noisy targets. We conclude that noise-free training data is unnecessary in this application.
That said, saturation (gamut clipping) renders the expectation incorrect due to removing part of the distribution. As saturation is unwanted for other reasons too, this is not a significant limitation.%

\textbf{Multiplicative Bernoulli noise}
(aka binomial noise) constructs a random mask $m$ that is 1 for valid pixels and 0 for zeroed/missing pixels. To avoid backpropagating gradients from missing pixels, we exclude them from the loss:%
\begin{equation}
\argmin_\theta \sum_i \left(m \odot (f_\theta(\inputx_i) - \inputy_i \right))^2, \label{eq:s2s}
\end{equation}
as described by Ulyanov et al. \yrcite{Ulyanov2017b} in the context of their deep image prior (DIP). 

The probability of corrupted pixels is denoted with $p$; in our training we vary $p \in [0.0, 0.95]$ and during testing $p=0.5$. Training with clean targets gives an average of 31.85\ppm0.03\,dB, noisy targets (separate $m$ for input and target) give a slightly higher 32.02\ppm0.03\,dB, possibly because noisy targets effectively implement a form of dropout \cite{srivastava2014} at the network output. 
DIP was almost 2\,dB worse -- DIP is not a learning-based solution, and as such very different from our approach, but it shares the property that neither clean examples nor an explicit model of the corruption is needed. We used the ``Image  reconstruction'' setup as described in the DIP supplemental material%
.\footnote{https://dmitryulyanov.github.io/deep\_image\_prior}

\figsimple
\figtexremoval
\figmoderecovery
\figsweep

\textbf{Text removal}
\label{sec:textremoval}
Figure~\ref{fig:textremoval} demonstrates blind text removal. The corruption consists of a large, varying number of random strings in random places, also on top of each other, and furthermore so that the font size and color are randomized as well. The font and string orientation remain fixed.

The network is trained using independently corrupted input and target pairs. The probability of corrupted pixels $p$ is approximately $[0,0.5]$ during training, and $p \approx 0.25$ during testing. In this test the mean ($L_2$ loss) is not the correct answer because the overlaid text has colors unrelated to the actual image, and the resulting image would incorrectly tend towards a linear combination of the right answer and the average text color (medium gray). However, with any reasonable amount of overlaid text, a pixel retains the original color more often than not, and therefore the median is the correct statistic. Hence, we use $L_1 = |f_\theta(\inputx)-\inputy|$ as the loss function. Figure~\ref{fig:textremoval} shows an example result.

\textbf{Random-valued impulse noise}
\label{sec:impulse}
replaces some pixels with noise and retains the colors of others. Instead of the standard salt and pepper noise (randomly replacing pixels with black or white), we study a harder distribution where each pixel is replaced with a random color drawn from the uniform distribution $[0,1]^3$ with probability $p$ and retains its color with probability $1-p$. The pixels' color distributions are a Dirac at the original color plus a uniform distribution, with relative weights given by the replacement probability $p$. In this case, neither the mean nor the median yield the correct result; the desired output is the \emph{mode} of the distribution (the Dirac spike). The distribution remains unimodal. For approximate mode seeking, we use an annealed version of the ``$L_0$ loss'' function defined as $(|f_\theta(\inputx)-\inputy|+\epsilon)^\gamma$, where $\epsilon = 10^{-8}$, where $\gamma$ is annealed linearly from $2$ to $0$ during training. This annealing did not cause any numerical issues in our tests. The relationship of the $L_0$ loss and mode seeking is analyzed in \thesupplement.

We again train the network using noisy inputs and noisy targets, where the probability of corrupted pixels is randomized separately for each pair from $[0, 0.95]$. Figure~\ref{fig:moderecovery} shows the inference results when 70\% input pixels are randomized. Training with $L_2$ loss biases the results heavily towards gray, because the result tends towards a linear combination the correct answer and and mean of the uniform random corruption. As predicted by theory, the $L_1$ loss gives good results as long as fewer than 50\% of the pixels are randomized, but beyond that threshold it quickly starts to bias dark and bright areas towards gray (Figure~\ref{fig:sweep}). $L_0$, on the other hand, shows little bias even with extreme corruptions (e.g. 90\% pixels), because of all the possible pixel values, the correct answer (e.g. 10\%) is still the most common.

\figmclossfuncs

\subsection{Monte Carlo Rendering}
\label{sec:mc}

Physically accurate renderings of virtual environments are most often generated through a process known as Monte Carlo path tracing. This amounts to drawing random sequences of scattering events (``light paths'') in the scene that connect light sources and virtual sensors, and integrating the radiance carried by them over all possible paths \cite{Veach1995}.
The Monte Carlo integrator is constructed such that the intensity of each pixel is the expectation of the random path sampling process, i.e., the sampling noise is zero-mean. However, despite decades of research into importance sampling techniques, little else can be said about the distribution. It varies from pixel to pixel, heavily depends on the scene configuration and rendering parameters, and can be arbitrarily multimodal. %
Some lighting effects, such as focused caustics, also result in extremely long-tailed distributions with rare, bright outliers.

All of these effects make the removal of Monte Carlo noise much more difficult than removing, e.g., Gaussian noise. On the other hand, the problem is somewhat alleviated by the possibility of generating auxiliary information that has been empirically found to correlate with the clean result during data generation. In our experiments, the denoiser input consists of not only the per-pixel luminance values, but also the average albedo (i.e., texture color) and normal vector of the surfaces visible at each pixel.

\textbf{High dynamic range (HDR)\ \ }
Even with adequate sampling, the floating-point pixel luminances may differ from each other by several orders of magnitude. %
In order to construct an image suitable for the generally 8-bit display devices, this high dynamic range needs to be compressed to a fixed range using a tone mapping operator \cite{ToneMapSurvey}.
We use a variant of Reinhard's global operator~\cite{Reinhard2002}: $T(v)=(v/(1+v))^{1/2.2}$, where $v$ is a scalar luminance value, possibly pre-scaled with an image-wide exposure constant. This operator maps any $v \ge 0$ into range $0 \le T(v) < 1$.

The combination of virtually unbounded range of luminances and the nonlinearity of operator $T$ poses a problem. If we attempt to train a denoiser that outputs luminance values $v$, a standard MSE loss \mbox{$L_2=(f_\theta(\inputx)-\inputy)^2$} will be dominated by the long-tail effects (outliers) in the targets, and training does not converge. On the other hand, if the denoiser were to output tonemapped values $T(v)$, the nonlinearity of $T$ would make the expected value of noisy target images $\expectation{T(v)}$ different from the clean training target $T(\expectation{v})$, leading to incorrect predictions.%

A metric often used for measuring the quality of HDR images is the relative MSE~\cite{Rousselle2011}, where the squared difference is divided by the square of approximate luminance of the pixel, i.e.,
\mbox{$(f_\theta(\inputx)-\inputy)^2 / (\inputy + \epsilon)^2$}. 
However, this metric suffers from the same nonlinearity problem as comparing of tonemapped outputs.
Therefore, we propose to use the network output, which tends towards the correct value in the limit, in the denominator:
\mbox{$L_\textrm{HDR}=(f_\theta(\inputx)-\inputy)^2 / (f_\theta(\inputx) + 0.01)^2$}.
It can be shown that $L_\textrm{HDR}$ converges to the correct expected value as long as we consider the gradient of the denominator to be zero.

Finally, we have observed that it is beneficial to tone map the input image $T(\inputx)$ instead of using HDR inputs. The network continues to output non-tonemapped (linear-scale) luminance values, retaining the correctness of the expected value. Figure~\ref{fig:mclossfuncs} evaluates the different loss functions.

\figmcresults

\textbf{Denoising Monte Carlo rendered images\ \ }
We trained a denoiser for Monte Carlo path traced images rendered using 64 samples per pixel (spp). Our training set consisted of 860~architectural images, and the validation was done using 34~images from a different set of scenes. Three versions of the training images were rendered: two with 64 spp using different random seeds (noisy input, noisy target), and one with 131k spp (clean target). The validation images were rendered in both 64 spp (input) and 131k spp (reference) versions. All images were 960$\times$540 pixels in size, and as mentioned earlier, we also saved the albedo and normal buffers for all of the input images. Even with such a small dataset, rendering the 131k~spp clean images was a strenuous effort --- for example, Figure~\ref{fig:mcresults}d took 40 minutes to render on a high-end graphics server with 8 $\times$ NVIDIA Tesla P100 GPUs and a 40-core Intel Xeon CPU.

The average PSNR of the 64 spp validation inputs with respect to the corresponding reference images was~22.31\,dB (see Figure~\ref{fig:mcresults}a for an example). The network trained for 2000~epochs using clean target images reached an average PSNR of~31.83\,dB on the validation set, whereas the similarly trained network using noisy target images gave 0.5\,dB less. Examples are shown in Figure~\ref{fig:mcresults}b,c -- the training took 12 hours with a single NVIDIA Tesla P100 GPU. 

At 4000 epochs, the noisy targets matched 31.83\,dB, i.e., noisy targets took approximately twice as long to converge. However, the gap between the two methods had not narrowed appreciably, leading us to believe that some quality difference will remain even in the limit. This is not surprising, since the training dataset contained only a limited number of training pairs (and thus noise realizations) due to the cost of generating the clean target images, and we wanted to test both methods using matching data.
That said, given that noisy targets are 2000 times faster to produce, one could trivially produce a larger quantity of them and still realize vast gains. The finite capture budget study (Section~\ref{sec:gaussian}) supports this hypothesis.

\figPSNRonline

\textbf{Online training\ \ }
Since it can be tedious to collect a sufficiently large corpus of Monte Carlo images for training a generally applicable denoiser, a possibility is to train a model specific to a single 3D scene, e.g., a game level or a movie shot \cite{Chaitanya2017}. In this context, it can even be desirable to train on-the-fly while walking through the scene. In order to maintain interactive frame rates, we can afford only few samples per pixel, and thus both input and target images will be inherently noisy.

Figure~\ref{fig:PSNRonline} shows the convergence plots for an experiment where we trained a denoiser from scratch for the duration of 1000 frames in a scene flythrough. On an NVIDIA Titan V GPU, path tracing a single 512$\times$512 pixel image with 8~spp took 190\,ms, and we rendered two images to act as input and target. A single network training iteration with a random 256$\times$256 pixel crop took 11.25\,ms and we performed eight of them per frame. Finally, we denoised both rendered images, each taking 15\,ms, and averaged the result to produce the final image shown to the user. Rendering, training and inference took 500\,ms/frame.

Figure~\ref{fig:PSNRonline} shows that training with clean targets does not perform appreciably better than noisy targets. As rendering a single clean image takes approx. 7~minutes in this scene (resp. 190\,ms for a noisy target), the quality/time tradeoff clearly favors noisy targets. 

\subsection{Magnetic Resonance Imaging (MRI)}
\label{sec:mri}
Magnetic Resonance Imaging (MRI) produces volumetric images of biological tissues essentially by %
sampling the Fourier transform (the ``$k$-space'') of the signal. Modern MRI techniques have long relied on compressed sensing (CS) to cheat the Nyquist-Shannon limit: they undersample $k$-space, and perform non-linear reconstruction that removes aliasing by exploiting the sparsity of the image in a suitable transform domain \cite{Lustig2007}.

We observe that if we turn the $k$-space sampling into a random process with a known probability density $p(k)$ over the frequencies $k$, our main idea applies. In particular, we model the $k$-space sampling operation as a Bernoulli process where each individual frequency has a probability $p(k) = e^{-\lambda |k|}$ of being selected for acquisition.\footnote{Our simplified example deviates from practical MRI in the sense that we do not sample the spectra along 1D trajectories. However, we believe that designing pulse sequences that lead to similar pseudo-random sampling characteristics is straightforward.} The frequencies that are retained are weighted by the inverse of the selection probability, and non-chosen frequencies are set to zero. Clearly, the expectation of this ``Russian roulette'' process is the correct spectrum. The parameter $\lambda$ controls the overall fraction of $k$-space retained; in the following experiments, we choose it so that $10\%$ of the samples are retained relative to a full Nyquist-Shannon sampling. The undersampled spectra are transformed to the primal image domain by the standard inverse Fourier transform. An example of an undersampled input/target picture, the corresponding fully sampled reference, and their spectra, are shown in Figure~\ref{fig:mriresults}(a, d).

Now we simply set up a regression problem of the form \eqref{eq:n2n} and train a convolutional neural network using pairs of two independent undersampled images $\inputx$ and $\inputy$ of the same volume. As the spectra of the input and target are correct on expectation, and the Fourier transform is linear, we use the $L_2$ loss. Additionally, we improve the result slightly by enforcing the exact preservation of frequencies that are present in the input image $\inputx$ by Fourier transforming the result $f_\theta(\inputx)$, replacing the frequencies with those from the input, and transforming back to the primal domain before computing the loss: the final loss reads $(\mathcal{F}^{-1} (R_{\inputx}(\mathcal{F}(f_\theta(\inputx)))) - \inputy)^2$, where $R$ denotes the replacement of non-zero frequencies from the input. This process is trained end-to-end.

We perform experiments on 2D slices extracted from the IXI brain scan MRI dataset.\footnote{http://brain-development.org/ixi-dataset $\to$ T1 images.} To simulate spectral sampling, we draw random samples from the FFT of the (already reconstructed) images in the dataset. Hence, in deviation from actual MRI samples, our data is real-valued and has the periodicity of the discrete FFT built-in. %
The training set contained 5000 images in 256$\times$256 resolution from 50~subjects, and for validation we chose 1000 random images from 10~different subjects.
The baseline PSNR of the sparsely-sampled input images was 20.03\,dB when reconstructed directly using IFFT. The network trained for 300 epochs with noisy targets reached an average PSNR of 31.74\,dB on the validation data, and the network trained with clean targets reached 31.77\,dB. Here the training with clean targets is similar to prior art \cite{Wang2016,Lee2017}. Training took 13~hours on an NVIDIA Tesla P100 GPU. Figure~\ref{fig:mriresults}(b, c) shows an example of reconstruction results between convolutional networks trained with noisy and clean targets, respectively. 
In terms of PSNR, our results quite closely match those reported in recent work. %

\figmriresults

\section{Discussion}

We have shown that simple statistical arguments lead to new capabilities in learned signal recovery using deep neural networks; it is possible to recover signals under complex corruptions \emph{without observing clean signals}, without an explicit statistical characterization of the noise or other corruption, at performance levels equal or close to using clean target data. That clean data is not necessary for denoising is not a new observation: indeed, consider, for instance, the classic BM3D algorithm \cite{Dabov2007} that draws on self-similar patches within a single noisy image. We show that the previously-demonstrated high restoration performance of deep neural networks can likewise be achieved entirely without clean data,
all based on the same general-purpose deep convolutional model.
This points the way to significant benefits in many applications by removing the need for potentially strenuous collection of clean data.

AmbientGAN \cite{ambientgan} trains generative adversarial networks \cite{Goodfellow2014} using corrupted observations. In contrast to our approach, AmbientGAN needs an explicit forward model of the corruption. We find combining ideas along both paths intriguing.

\section*{Acknowledgments}
Bill Dally, David Luebke, Aaron Lefohn for discussions and supporting the research; NVIDIA Research staff for suggestions and discussion; Runa Lober and Gunter Sprenger for synthetic off-line training data; Jacopo Pantaleoni for the interactive renderer used in on-line training; Samuli Vuorinen for initial photography test data; Koos Zevenhoven for discussions on MRI; Peyman Milanfar for helpful comments.

\bibliography{paper}
\bibliographystyle{icml2018}

\ifarxiv
\clearpage
\renewcommand{\floatpagefraction}{.99}
\renewcommand{\topfraction}{.99}
\renewcommand{\bottomfraction}{.99}
\renewcommand{\textfraction}{.01}
\appendix
\section{Appendix}
\newcommand{\appendixsection}[1]{\subsection{#1}}
\input{appendix-content}

\fi

\end{document}

%% file: figures.tex
\newcommand{\h}{0mm}
\newcommand{\hh}{0mm}
\newcommand{\hhh}{0mm}

\newcommand{\figsimple}{
\renewcommand{\h}{0.236\linewidth}
\renewcommand{\hh}{0.118\linewidth}
\renewcommand{\hhh}{2.45mm}
\begin{figure}[t]
\centering
\makebox[15mm][c]{\small\textbf{(a) Gaussian ($\sigma=25$)}}\\
\includegraphics[width=\h]{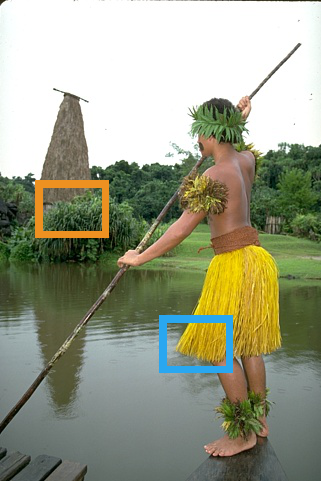}\hfill%
\includegraphics[width=\h]{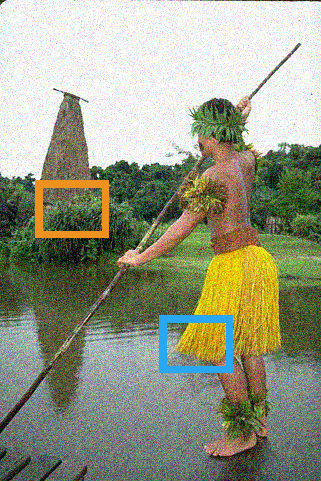}\hfill%
\includegraphics[width=\h]{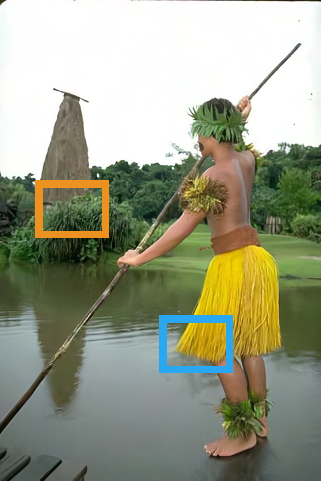}\hfill%
\includegraphics[width=\h]{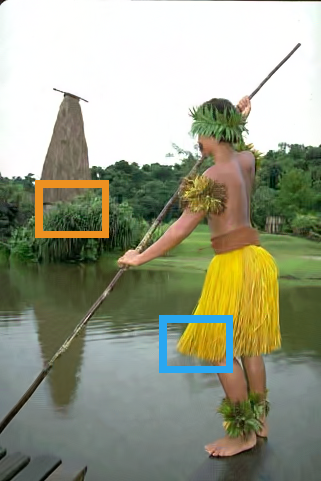}\hfill%
\rotatebox{90}{\makebox[20mm][c]{\small{BM3D}}}\\
\includegraphics[width=\hh]{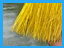}%
\includegraphics[width=\hh]{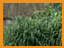}\hfill%
\includegraphics[width=\hh]{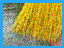}%
\includegraphics[width=\hh]{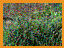}\hfill%
\includegraphics[width=\hh]{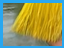}%
\includegraphics[width=\hh]{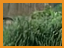}\hfill%
\includegraphics[width=\hh]{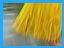}%
\includegraphics[width=\hh]{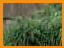}\hfill%
\makebox[\hhh][c]{}\\
\makebox[15mm][c]{\small\textbf{(b) Poisson ($\lambda=30$)}}\\
\includegraphics[width=\h]{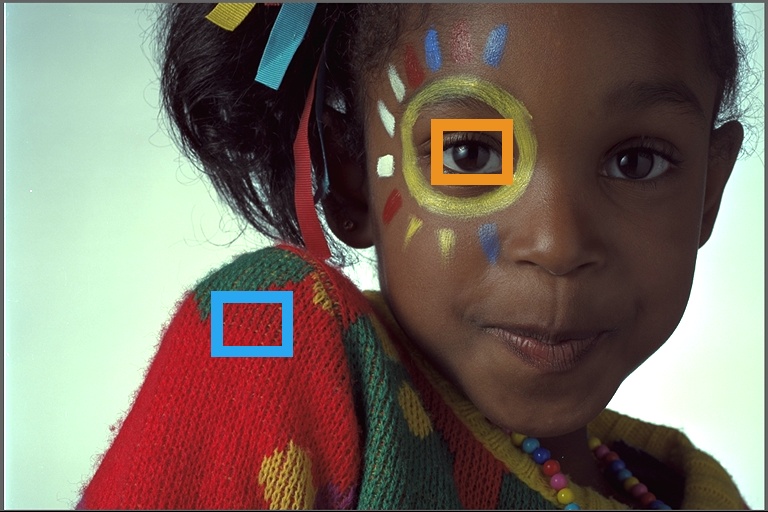}\hfill%
\includegraphics[width=\h]{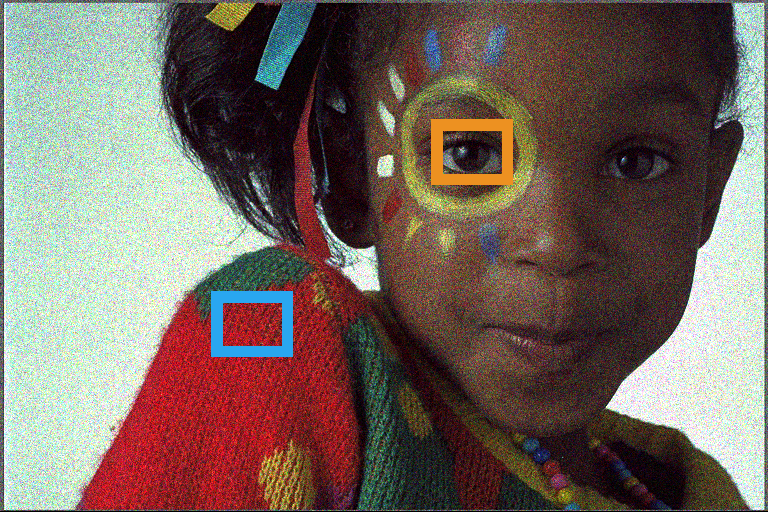}\hfill%
\includegraphics[width=\h]{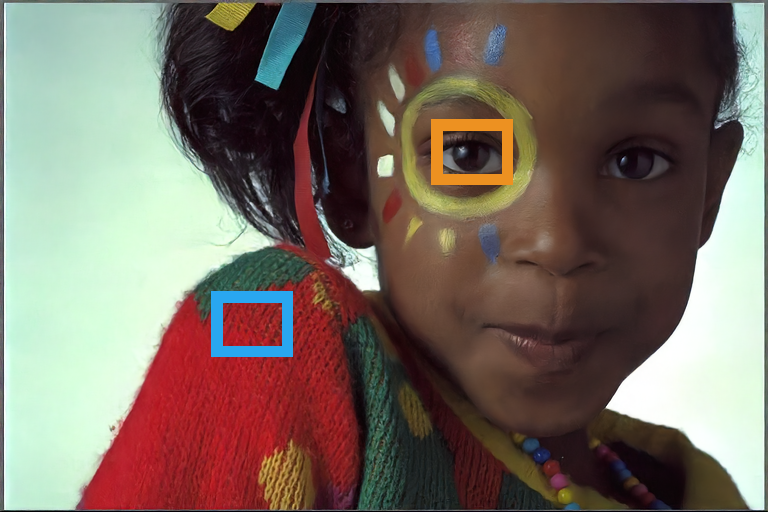}\hfill%
\includegraphics[width=\h]{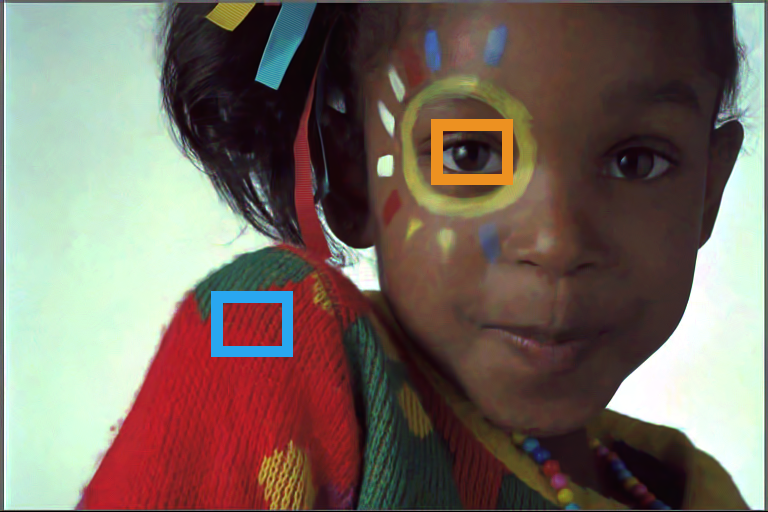}\hfill%
\rotatebox{90}{\makebox[4mm][c]{\small{\textsc{Anscombe}}}}\\
\includegraphics[width=\hh]{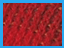}%
\includegraphics[width=\hh]{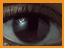}\hfill%
\includegraphics[width=\hh]{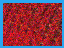}%
\includegraphics[width=\hh]{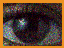}\hfill%
\includegraphics[width=\hh]{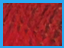}%
\includegraphics[width=\hh]{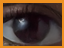}\hfill%
\includegraphics[width=\hh]{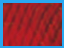}%
\includegraphics[width=\hh]{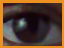}\hfill%
\makebox[\hhh][c]{}\\
\makebox[15mm][c]{\small\textbf{(c) Bernoulli ($p=0.5$)}}\\
\includegraphics[width=\h]{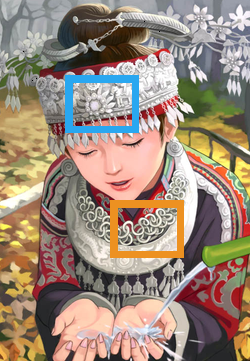}\hfill%
\includegraphics[width=\h]{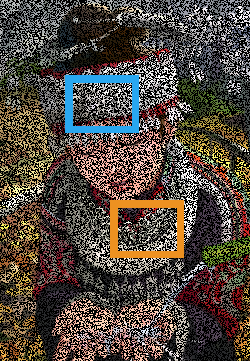}\hfill%
\includegraphics[width=\h]{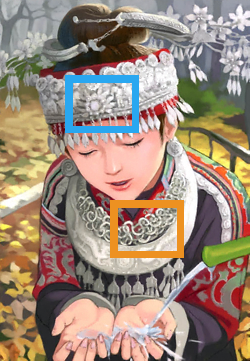}\hfill%
\includegraphics[width=\h]{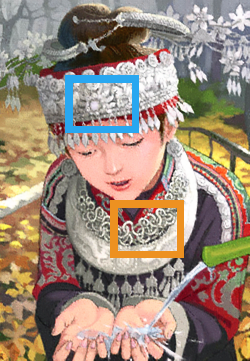}\hfill%
\rotatebox{90}{\makebox[20mm][c]{\small{\textsc{Deep Image Prior}}}}\\
\includegraphics[width=\hh]{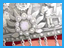}%
\includegraphics[width=\hh]{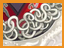}\hfill%
\includegraphics[width=\hh]{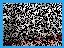}%
\includegraphics[width=\hh]{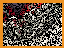}\hfill%
\includegraphics[width=\hh]{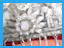}%
\includegraphics[width=\hh]{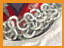}\hfill%
\includegraphics[width=\hh]{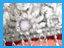}%
\includegraphics[width=\hh]{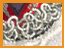}\hfill%
\makebox[\hhh][c]{}\\
\makebox[\h][c]{\small Ground truth}\hfill%
\makebox[\h][c]{\small Input}\hfill%
\makebox[\h][c]{\small Our}\hfill%
\makebox[\h][c]{\small Comparison}\hfill%
\makebox[\hhh][c]{}\\
\caption{Example results for Gaussian, Poisson, and Bernoulli noise. Our result was computed by using noisy targets --- the corresponding result with clean targets is omitted because it is virtually identical in all three cases, as discussed in the text. A different comparison method is used for each noise type.
}
\label{fig:simple}
\end{figure}
}

\newcommand{\figmriresults}{
\renewcommand{\h}{0.229\linewidth}
\renewcommand{\hh}{\hspace*{1mm}}
\begin{figure}[t]
\centering
\begin{tabular}{@{}c@{\hh}c@{\hh}c@{\hh}c@{\hh}c@{}}
\rotatebox{90}{\makebox[\h][c]{\centering Image}}&
\includegraphics[width=\h]{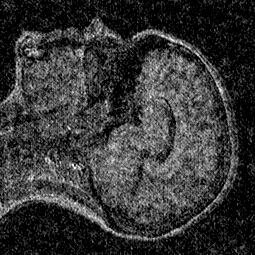}&
\includegraphics[width=\h]{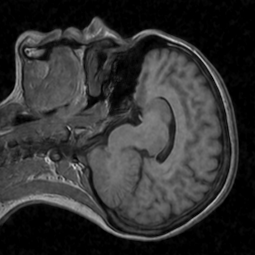}&
\includegraphics[width=\h]{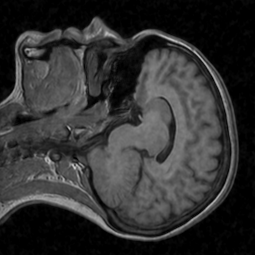}&
\includegraphics[width=\h]{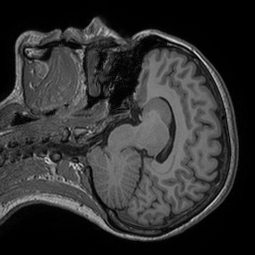}\vspace*{-.2mm}\\
\rotatebox{90}{\makebox[\h][c]{\centering Spectrum}}&
\includegraphics[width=\h]{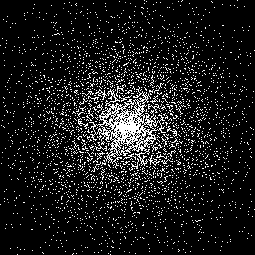}&
\includegraphics[width=\h]{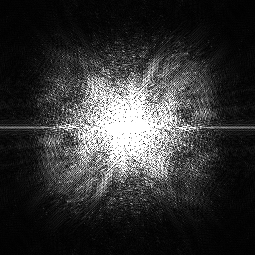}&
\includegraphics[width=\h]{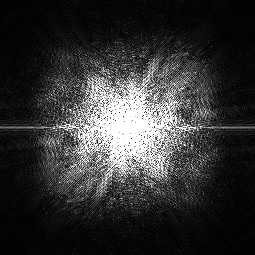}&
\includegraphics[width=\h]{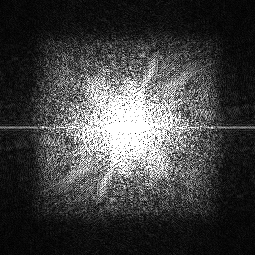}\\
&%
\small (a) Input&
\small (b) Noisy trg.&
\small (c) Clean trg.&
\small (d) Reference\\
&%
\small 18.93\,dB&
\small 29.77\,dB&
\small 29.81\,dB&
\\
\end{tabular}
\caption{\label{fig:mriresults}%
MRI reconstruction example.
(a)~Input image with only 10\% of spectrum samples retained and scaled by $1/p$.
(b)~Reconstruction by a network trained with noisy target images similar to the input image.
(c)~Same as previous, but training done with clean target images similar to the reference image.
(d)~Original, uncorrupted image.
PSNR values refer to the images shown here, see text for averages over the entire validation set.
}
\end{figure}
}

\newcommand{\figconvergence}{
\renewcommand{\h}{0.32\linewidth}
\renewcommand{\hh}{0.08\linewidth}
\begin{figure*}[t]
\centering
\includegraphics[width=\h,trim={22mm 22mm 22mm 22mm},clip]{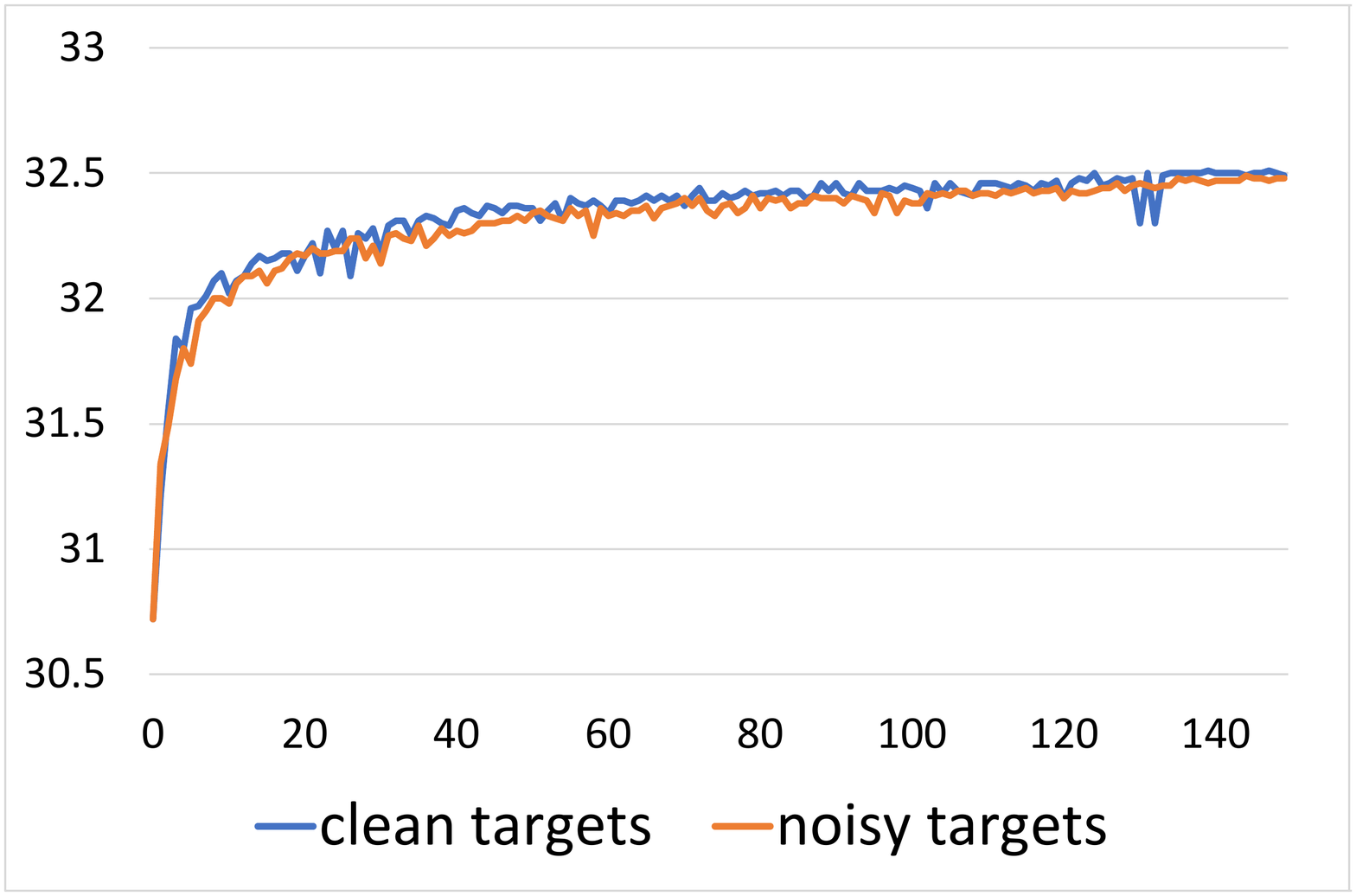}%
\raisebox{9mm}{\makebox[0mm][l]{\hspace*{-18mm}\includegraphics[width=\hh,trim={0 0 0 0},clip]{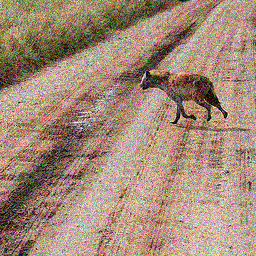}}}%
\hfill%
\includegraphics[width=\h,trim={22mm 22mm 22mm 22mm},clip]{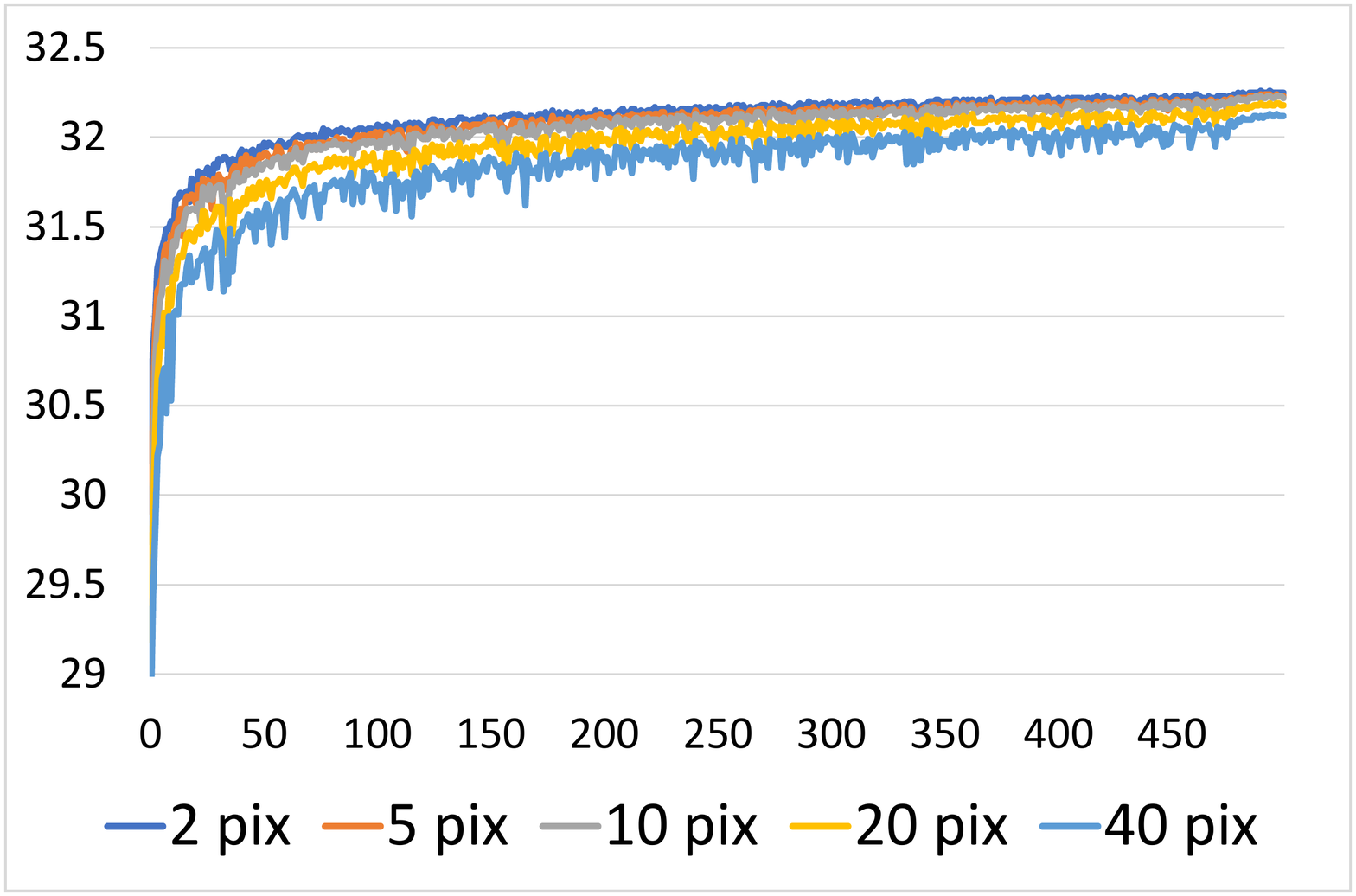}%
\raisebox{9mm}{\makebox[0mm][l]{\hspace*{-18mm}\includegraphics[width=\hh,trim={0 0 0 0},clip]{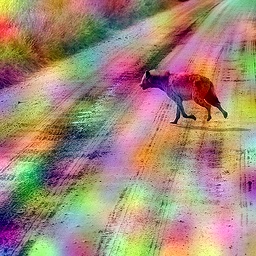}}}%
\hfill
\includegraphics[width=\h,trim={22mm 22mm 22mm 22mm},clip]{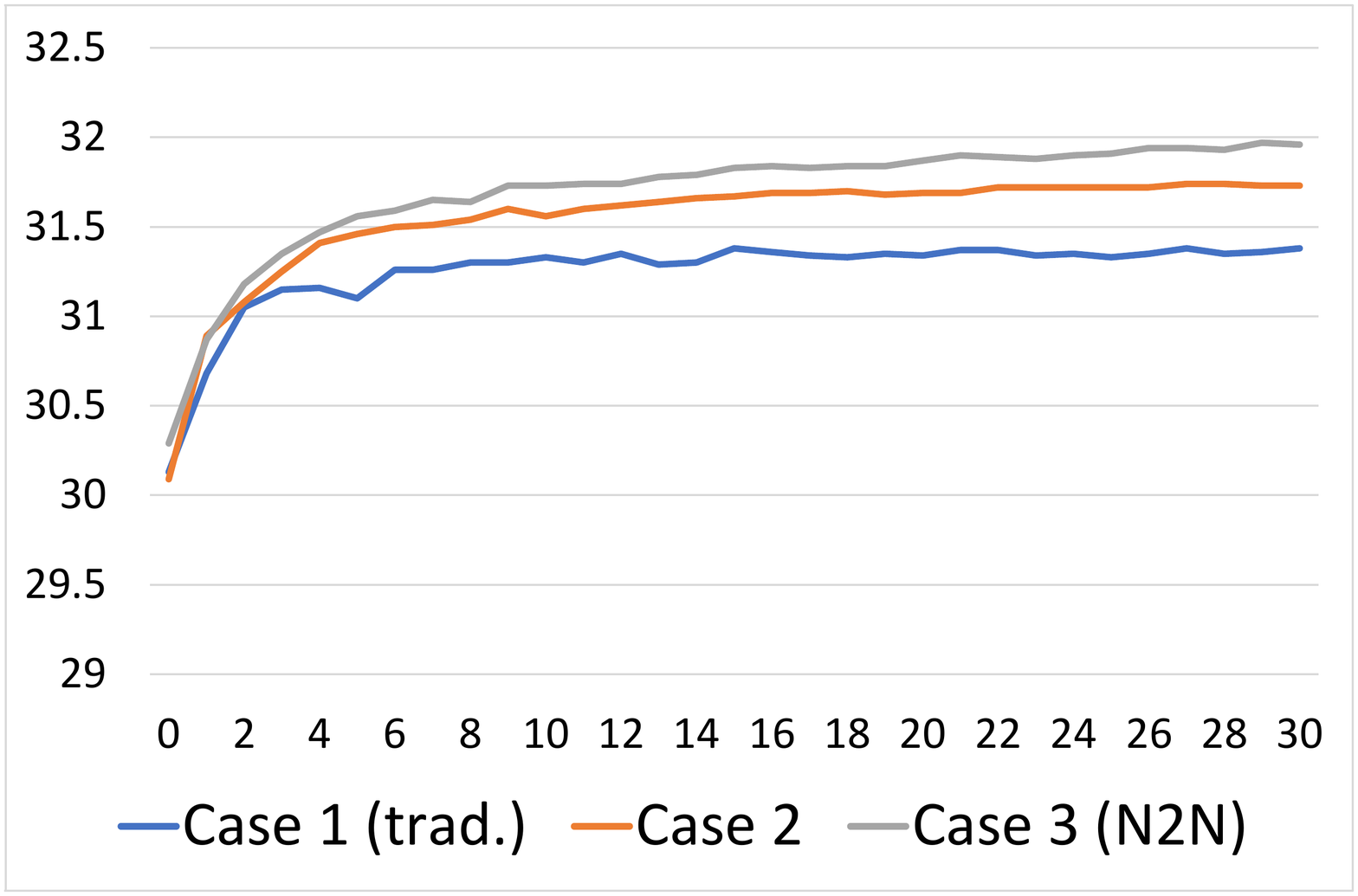}\\
\makebox[\h][c]{\footnotesize\textbf{(a) White Gaussian, $\sigma = 25$}}\hfill
\makebox[\h][c]{\footnotesize\textbf{(b) Brown Gaussian, $\sigma = 25$}}\hfill
\makebox[\h][c]{\footnotesize\textbf{(c) Capture budget study (see text)}}\\
\caption{Denoising performance ($dB$ in \Kodak~dataset) as a function of training epoch for additive Gaussian noise. 
(a) For i.i.d. (white) Gaussian noise, clean and noisy targets lead to very similar convergence speed and eventual quality.
(b) For brown Gaussian noise, we observe that increased inter-pixel noise correlation (wider spatial blur; one graph per bandwidth) slows convergence down, but eventual performance remains close.
(c) Effect of different allocations of a fixed capture budget to noisy vs. clean examples (see text).
}
\label{fig:convergence}
\end{figure*}
}

\newcommand{\tabpsnr}{
\begin{table}[t]
\centering
\caption{PSNR results from three test datasets \Kodak, \BSD, and \SET~for Gaussian, Poisson, and Bernoulli noise. The comparison methods are BM3D, Inverse Anscombe transform (ANSC), and deep image prior (DIP).}
\includegraphics[width=\columnwidth,trim={36mm 240mm 45mm 22mm},clip]{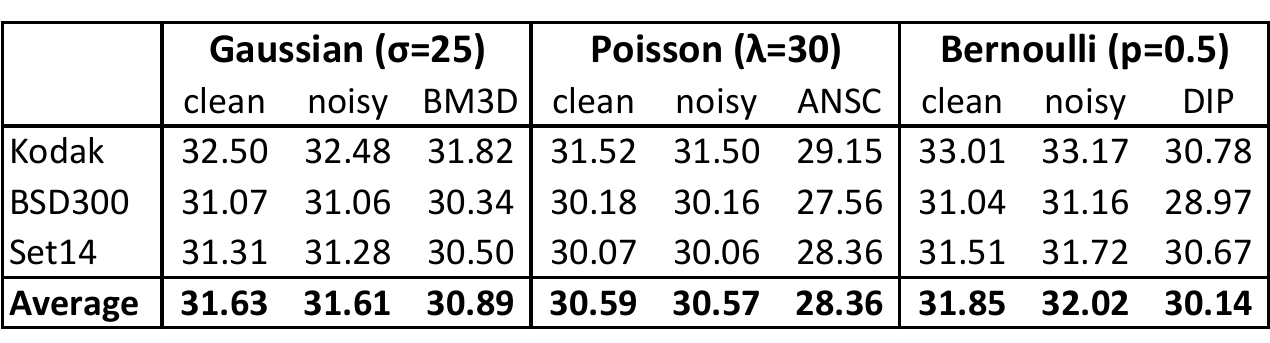}
\label{tab:psnr}
\end{table}
}

\newcommand{\figtexremoval}{
\renewcommand{\h}{0.16\linewidth}
\renewcommand{\hh}{0.0885\linewidth}
\begin{figure*}[t]
\centering
\raisebox{32mm}{\makebox[0mm][l]{\hspace{2mm}\small $p\approx 0.04$}}%
\raisebox{\hh+0.5mm}{\makebox[0mm][l]{\includegraphics[width=\hh]{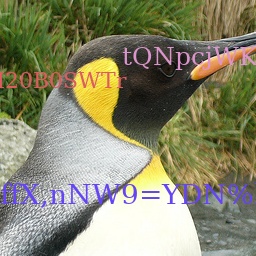}}}%
\includegraphics[width=\hh]{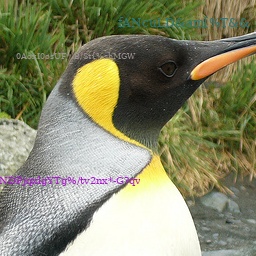}\hfill%
\raisebox{32mm}{\makebox[0mm][l]{\hspace{2mm}\small $p\approx 0.42$}}%
\raisebox{\hh+0.5mm}{\makebox[0mm][l]{\includegraphics[width=\hh]{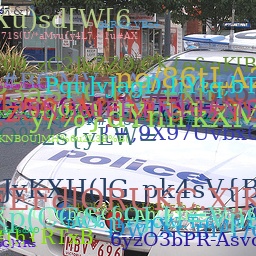}}}%
\includegraphics[width=\hh]{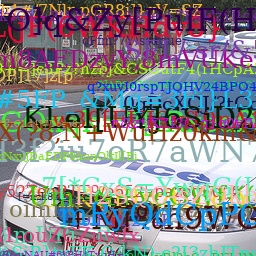}\hfill\hfill%
\includegraphics[width=\h]{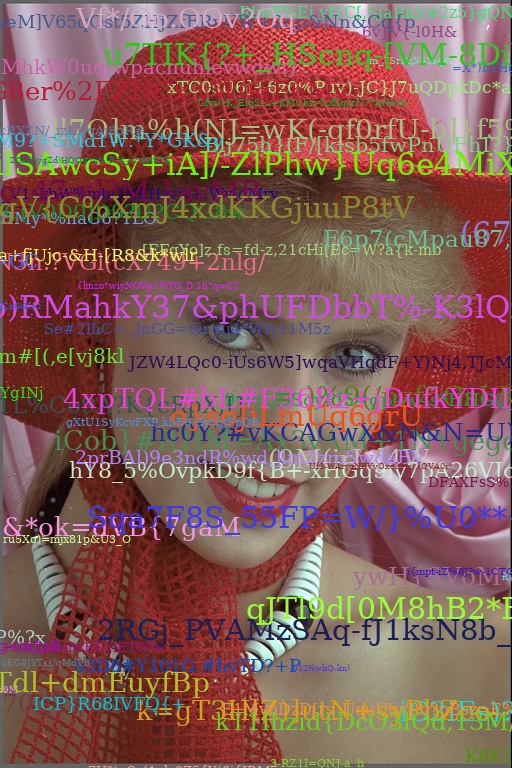}\hfill%
\includegraphics[width=\h]{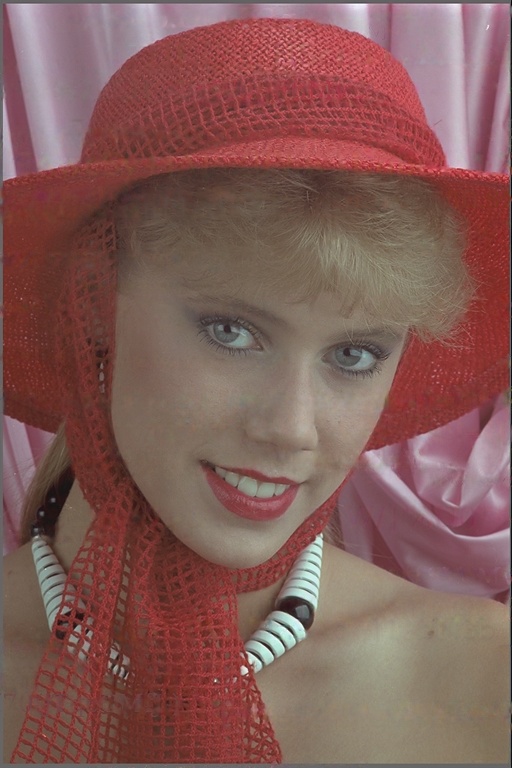}\hfill%
\includegraphics[width=\h]{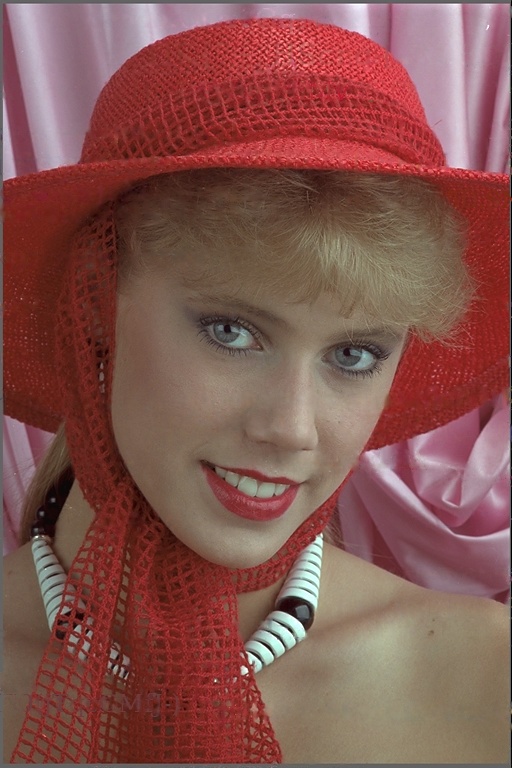}\hfill%
\includegraphics[width=\h]{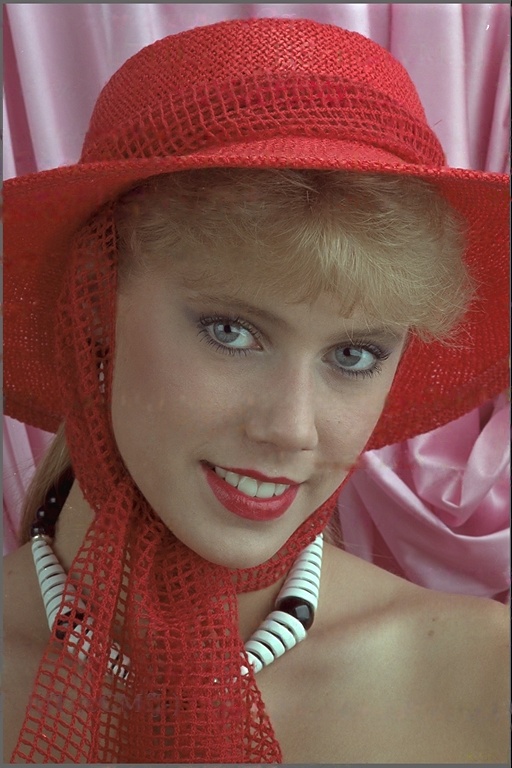}\hfill%
\includegraphics[width=\h]{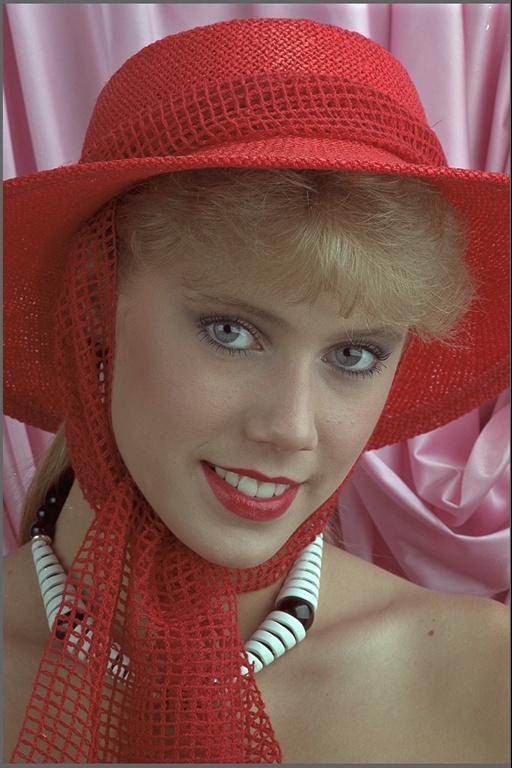}\\
\makebox[0.18\linewidth][c]{\small Example training pairs}\hfill%
\makebox[\h][c]{\small Input ($p\approx0.25$)}\hfill%
\makebox[\h][c]{\small $L_2$}\hfill%
\makebox[\h][c]{\small $L_1$}\hfill%
\makebox[\h][c]{\small Clean targets}\hfill%
\makebox[\h][c]{\small Ground truth}\\
\makebox[0.18\linewidth][c]{}\hfill%
\makebox[\h][c]{\small 17.12\,dB}\hfill%
\makebox[\h][c]{\small 26.89\,dB}\hfill%
\makebox[\h][c]{\small 35.75\,dB}\hfill%
\makebox[\h][c]{\small 35.82\,dB}\hfill%
\makebox[\h][c]{PSNR}\\
\caption{Removing random text overlays corresponds to seeking the median pixel color, accomplished using the $L_1$ loss. The mean ($L_2$ loss) is not the correct answer: note shift towards mean text color. Only corrupted images shown during training.}
\label{fig:textremoval}
\end{figure*}
}

\newcommand{\figmoderecovery}{
\renewcommand{\h}{0.16\linewidth}
\renewcommand{\hh}{0.0885\linewidth}
\begin{figure*}[t]
\centering
\raisebox{32mm}{\makebox[0mm][l]{\hspace{2mm}\small $p=0.22$}}%
\raisebox{\hh+0.5mm}{\makebox[0mm][l]{\includegraphics[width=\hh]{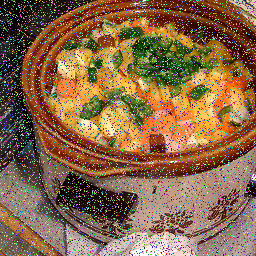}}}%
\includegraphics[width=\hh]{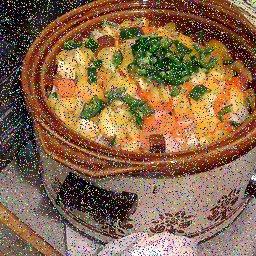}\hfill%
\raisebox{32mm}{\makebox[0mm][l]{\hspace{2mm}\small $p=0.81$}}%
\raisebox{\hh+0.5mm}{\makebox[0mm][l]{\includegraphics[width=\hh]{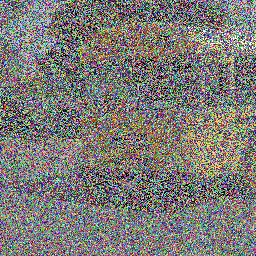}}}%
\includegraphics[width=\hh]{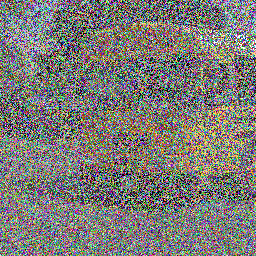}\hfill\hfill%
\includegraphics[width=\h]{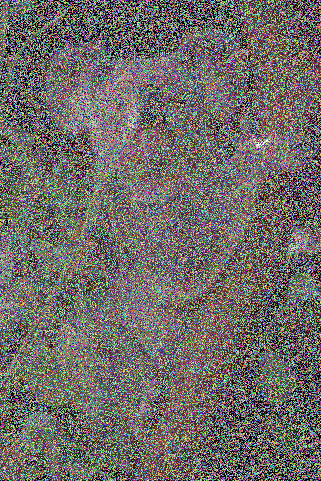}\hfill%
\includegraphics[width=\h]{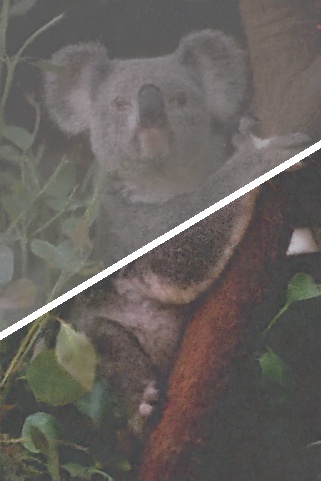}\hfill%
\includegraphics[width=\h]{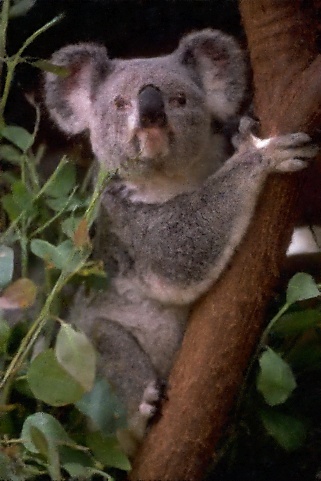}\hfill%
\includegraphics[width=\h]{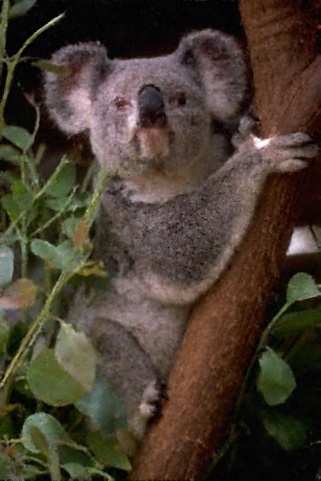}\hfill%
\includegraphics[width=\h]{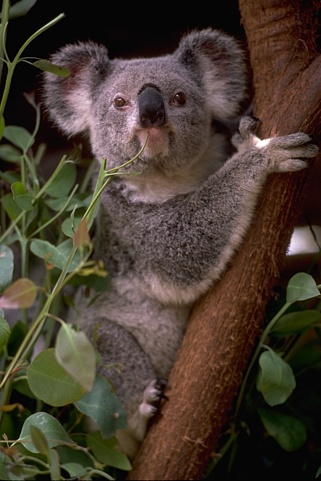}\\
\makebox[0.18\linewidth][c]{\small Example training pairs}\hfill%
\makebox[\h][c]{\small Input ($p=0.70$)}\hfill%
\makebox[\h][c]{\small $L_2$ / $L_1$}\hfill%
\makebox[\h][c]{\small $L_0$}\hfill%
\makebox[\h][c]{\small Clean targets}\hfill%
\makebox[\h][c]{\small Ground truth}\\
\makebox[0.18\linewidth][c]{}\hfill%
\makebox[\h][c]{\small 8.89\,dB}\hfill%
\makebox[\h][c]{\small 13.02\,dB / 16.36\,dB}\hfill%
\makebox[\h][c]{\small 28.43\,dB}\hfill%
\makebox[\h][c]{\small 28.86\,dB}\hfill%
\makebox[\h][c]{\small PSNR}\\
\caption{For random impulse noise, the approx. mode-seeking $L_0$ loss performs better than the mean ($L_2$) or median ($L_1$) seeking losses.}
\label{fig:moderecovery}
\end{figure*}
}

\newcommand{\figPSNRonline}{
\renewcommand{\h}{0.495\linewidth}
\begin{figure}[t]
\centering
\includegraphics[width=\linewidth,trim={20mm 55mm 20mm 68mm},clip]{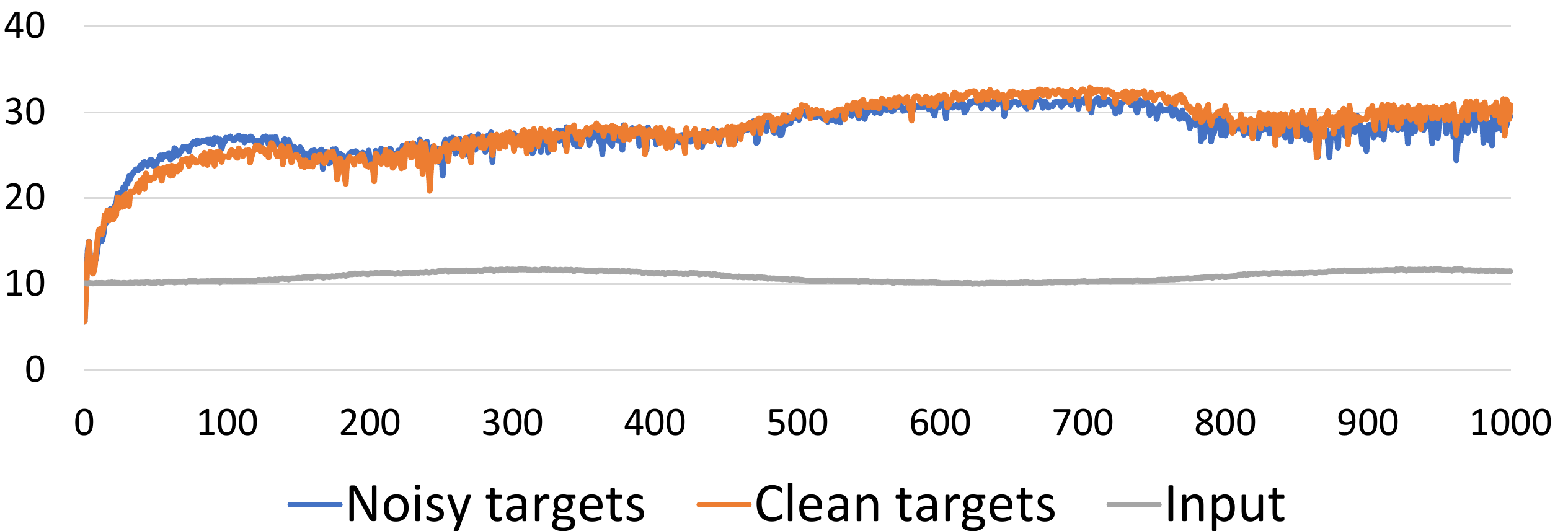}\\
\caption{Online training PSNR during a 1000-frame flythrough of the scene in Figure~\ref{fig:mclossfuncs}. Noisy target images are almost as good for learning as clean targets, but are over 2000$\times$  faster to render (190\,milliseconds vs 7\,minutes per frame in this scene). Both denoisers offer a substantial improvement over the noisy input.}
\label{fig:PSNRonline}
\end{figure}
}

\newcommand{\figsweep}{
\renewcommand{\h}{1.0\linewidth}
\begin{figure}[t]
\centering
\includegraphics[width=\h,trim={20mm 72mm 24mm 75mm},clip]{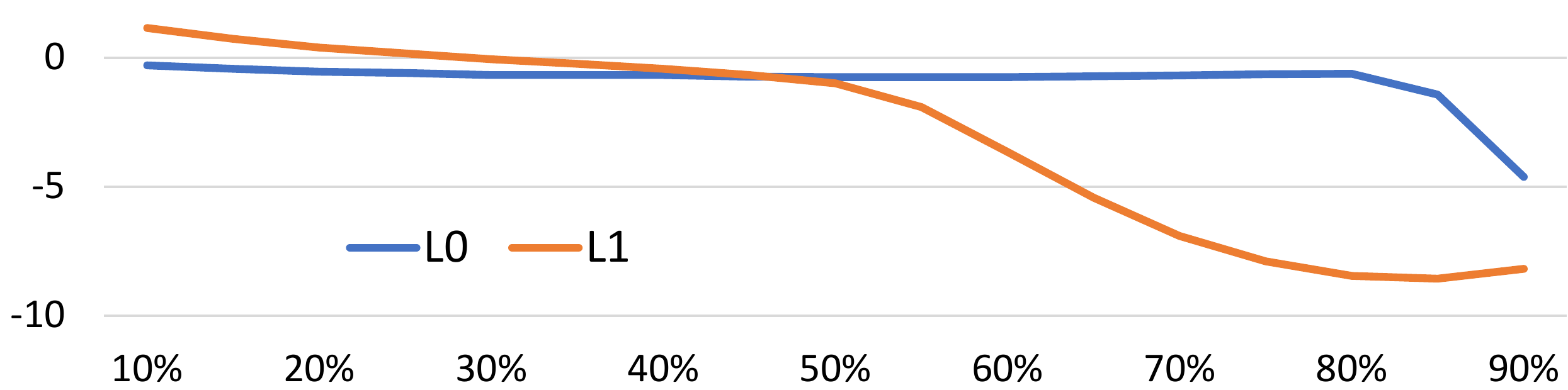}\\
\caption{PSNR of noisy-target training relative to clean targets with a varying percentage of target pixels corrupted by RGB impulse noise. In this test a separate network was trained for each corruption level, and the graph was averaged over the \Kodak~dataset.
}
\label{fig:sweep}
\end{figure}
}

\newcommand{\figmclossfuncs}{
\begin{figure*}[tb]
\setlength{\tabcolsep}{1pt}
\begin{tabular}{ccccccc}
	\includegraphics[width=0.139\linewidth]{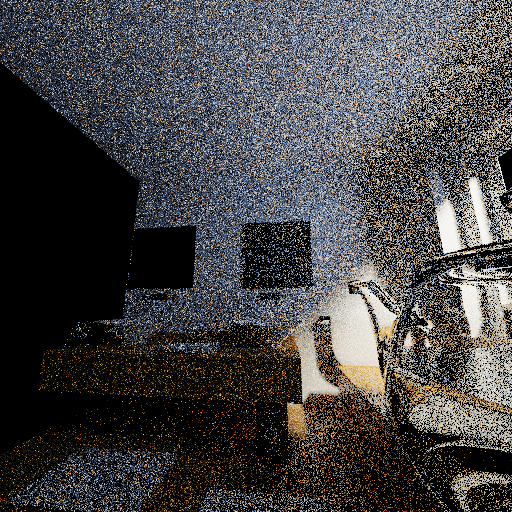} &%
	\includegraphics[width=0.139\linewidth]{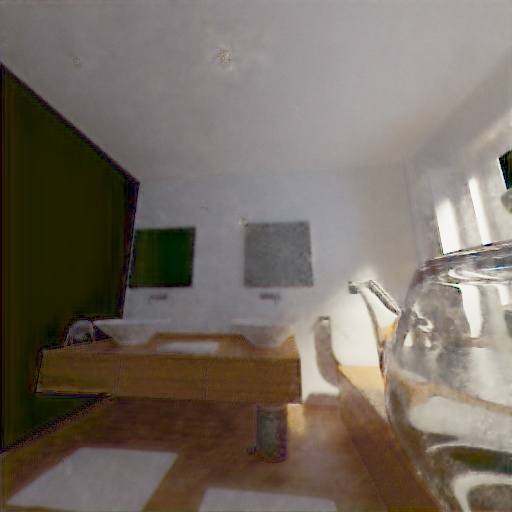} &
	\includegraphics[width=0.139\linewidth]{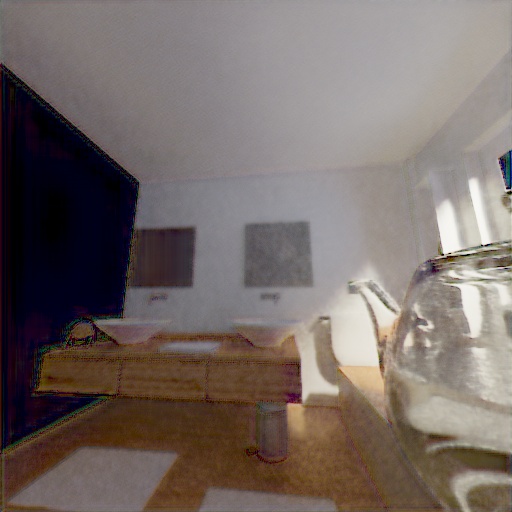} &
	\includegraphics[width=0.139\linewidth]{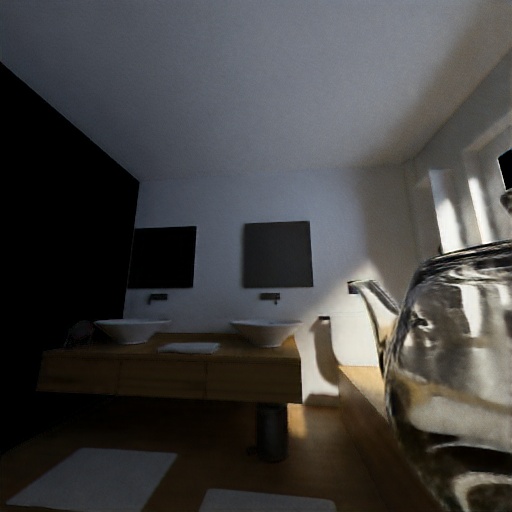} &
	\includegraphics[width=0.139\linewidth]{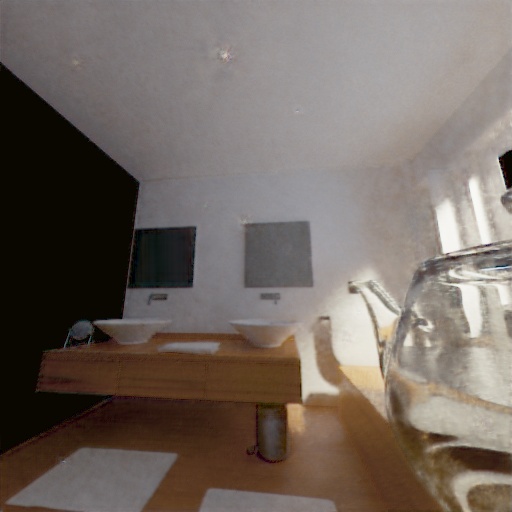} &
	\includegraphics[width=0.139\linewidth]{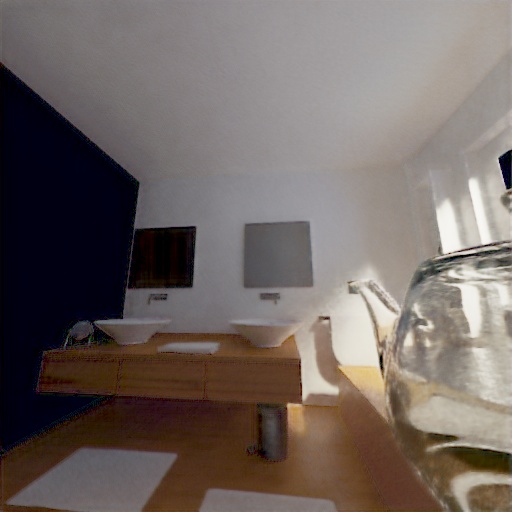} &
	\includegraphics[width=0.139\linewidth]{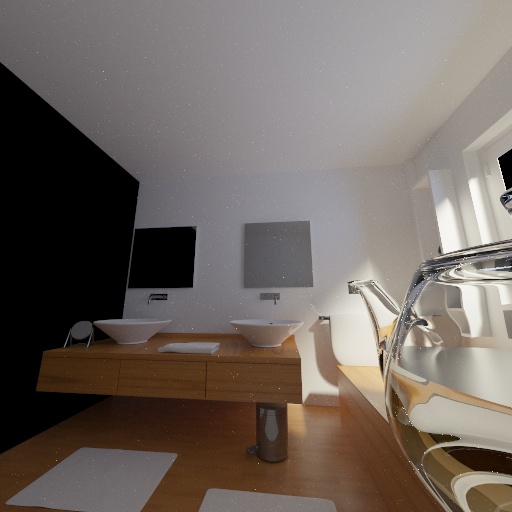} \\
	\small Input, 8 spp & 
	\small $L_2$ with $\inputx, \inputy$ &
	\small $L_2$ with $T(\inputx), \inputy$ &
	\small $L_2$ with $T(\inputx), T(\inputy)$ & 
	\small $L_\textrm{HDR}$ with $\inputx, \inputy$ &
	\small $L_\textrm{HDR}$ with $T(\inputx), \inputy$ & 
	\small Reference, 32k spp \\
	\small 11.32\,dB & 
	\small 25.46\,dB &
	\small 25.39\,dB &
	\small 15.50\,dB & 
	\small 29.05\,dB &
	\small 30.09\,dB & 
	\small PSNR\\
\end{tabular}%
	\caption{\label{fig:mclossfuncs}%
	Comparison of various loss functions for training a Monte Carlo denoiser with noisy target images rendered at 8 samples per pixel (spp). 
	In this high-dynamic range setting, our custom relative loss $L_\textrm{HDR}$ is clearly superior to $L_2$. Applying a non-linear tone map to the inputs is beneficial, while applying it to the target images skews the distribution of noise and leads to wrong, visibly too dark results.}	
\end{figure*}
}

\newcommand{\figmcresults}{
\renewcommand{\h}{0.2470\linewidth}
\renewcommand{\hh}{0.1235\linewidth}
\begin{figure*}[tb]
\includegraphics[width=\h]{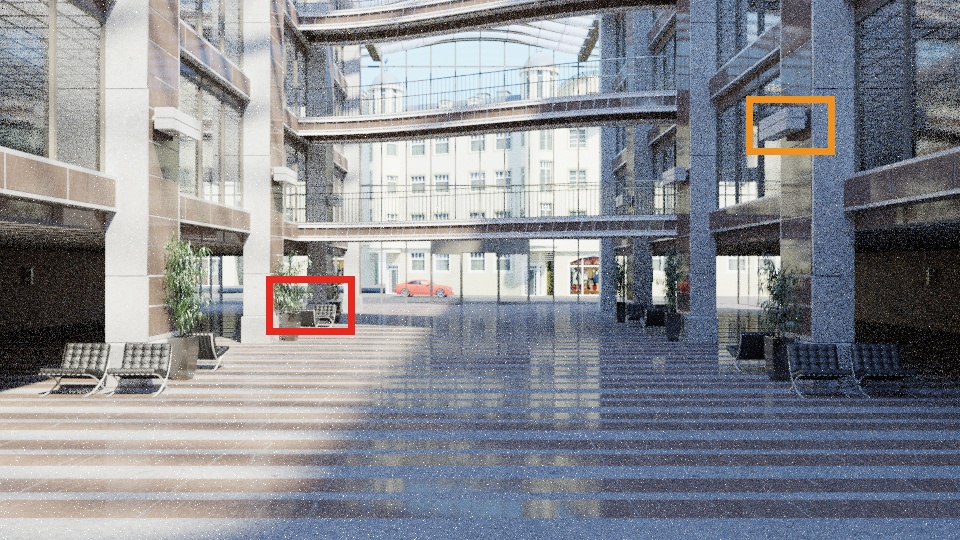}\hfill%
\includegraphics[width=\h]{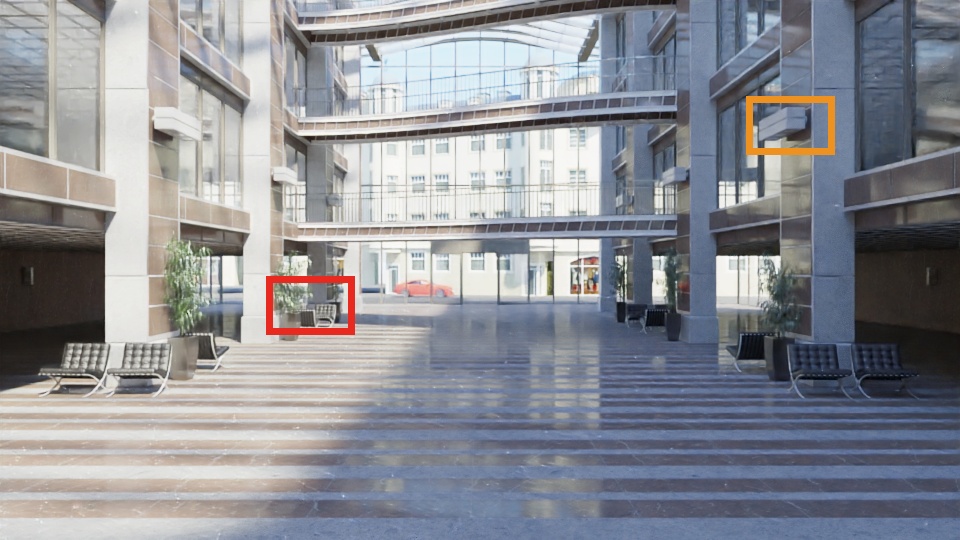}\hfill%
\includegraphics[width=\h]{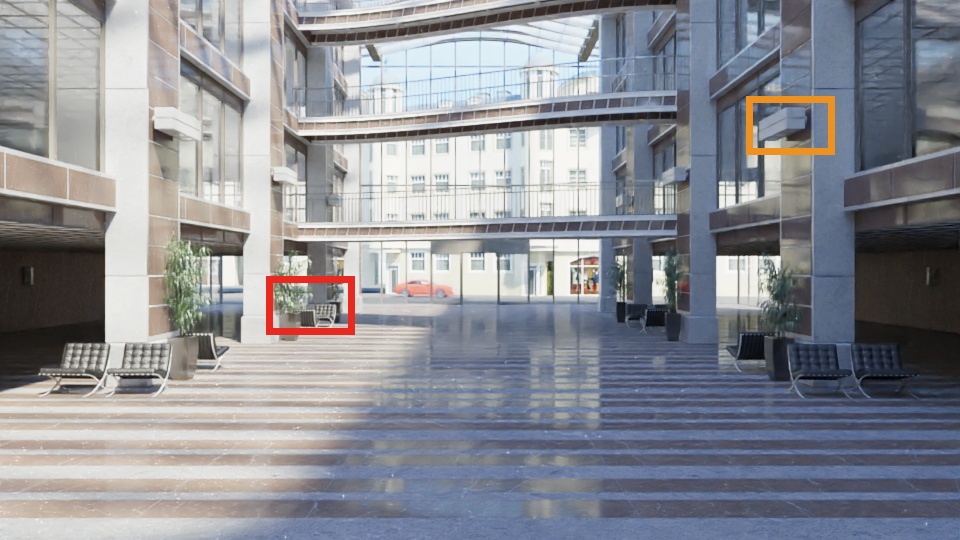}\hfill%
\includegraphics[width=\h]{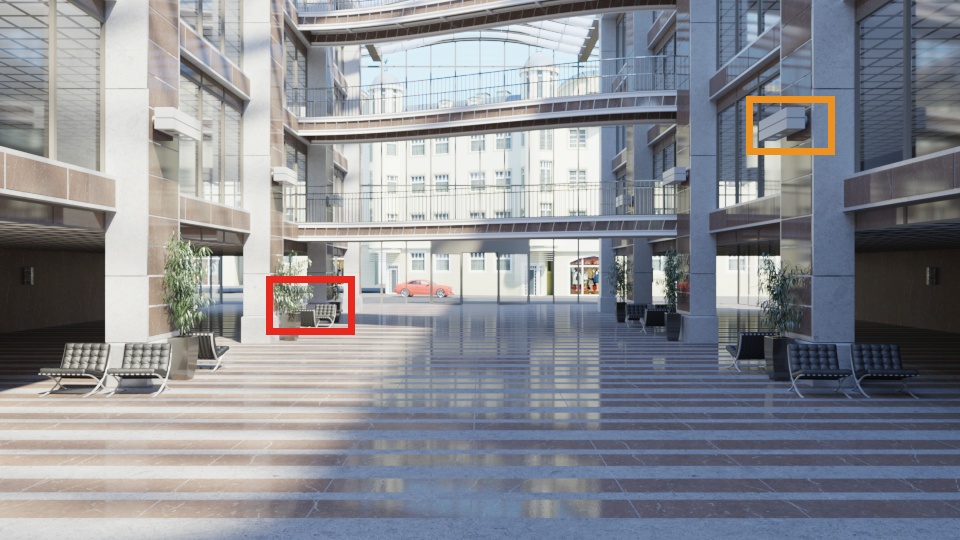}\\
\includegraphics[width=\hh]{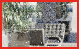}%
\includegraphics[width=\hh]{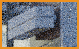}\hfill%
\includegraphics[width=\hh]{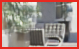}%
\includegraphics[width=\hh]{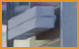}\hfill%
\includegraphics[width=\hh]{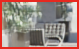}%
\includegraphics[width=\hh]{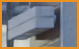}\hfill%
\includegraphics[width=\hh]{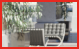}%
\includegraphics[width=\hh]{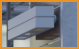}\hfill%
\makebox[\h][c]{\small (a) Input (64 spp), 23.93\,dB}\hfill%
\makebox[\h][c]{\small (b) Noisy targets, 32.42\,dB}\hfill%
\makebox[\h][c]{\small (c) Clean targets, 32.95\,dB}\hfill%
\makebox[\h][c]{\small (d) Reference (131k spp)}%
\caption{%
Denoising a Monte Carlo rendered image.
(a)~Image rendered with 64 samples per pixel.
(b)~Denoised 64 spp input, trained using 64 spp targets.
(c)~Same as previous, but trained on clean targets.
(d)~Reference image rendered with 131\,072 samples per pixel.
PSNR values refer to the images shown here, see text for averages over the entire validation set.
}
\label{fig:mcresults}
\end{figure*}
}

%% file: intro.tex
Signal reconstruction from corrupted or incomplete measurements is an important subfield of statistical data analysis. Recent advances in deep neural networks have sparked significant interest in avoiding the traditional, explicit a priori statistical modeling of signal corruptions, and instead \emph{learning} to map corrupted observations to the unobserved clean versions. This happens by training a regression model, e.g., a convolutional neural network (CNN), with a large number of pairs $(\inputx_i, y_i)$ of corrupted inputs $\inputx_i$ and clean targets $y_i$ and minimizing the empirical risk
\begin{equation}
\argmin_\theta \sum_i L\left( f_\theta(\inputx_i),\, y_i \right), \label{eq:regression}
\end{equation}
where $f_\theta$ is a parametric family of mappings (e.g., CNNs), under the loss function $L$. We use the notation $\inputx$ to underline the fact that the corrupted input $\inputx \sim p(\inputx | y_i)$ is a random variable distributed according to the clean target.
Training data may include, for example, pairs of short and long exposure photographs of the same scene, incomplete and complete k-space samplings of magnetic resonance images, fast-but-noisy and slow-but-converged ray-traced renderings of a synthetic scene, etc. Significant advances have been reported in several applications, including Gaussian denoising, de-JPEG, text removal \cite{Mao2016b}, super-resolution \cite{Ledig2016}, colorization \cite{Zhang2016colorization}, and image inpainting \cite{Iizuka2017}. Yet, obtaining clean training targets is often difficult or tedious: a noise-free photograph requires a long exposure; full MRI sampling precludes dynamic subjects; etc.

In this work, we observe that we can often \emph{learn to turn bad images into good images by only looking at bad images}, and do this just as well -- sometimes even better -- as if we were using clean examples. Further, we require neither an explicit statistical likelihood model of the corruption nor an image prior, and instead learn these indirectly from the training data. (Indeed, in one of our examples, synthetic Monte Carlo renderings, the non-stationary noise cannot be characterized analytically.) In addition to denoising, our observation is directly applicable to inverse problems such as MRI reconstruction from undersampled data.
While our conclusion is almost trivial from a statistical perspective, it significantly eases practical learned signal reconstruction by lifting requirements on availability of training data. %

The reference TensorFlow implementation for Noise2Noise training is available on GitHub.\footnote{\url{https://github.com/NVlabs/noise2noise}}

\section{Theoretical Background}
Assume that we have a set of unreliable measurements $(y_1, y_2, ...)$ of the room temperature. A common strategy for estimating the true unknown temperature is to find a number $z$ that has the smallest average deviation from the measurements according to some loss function $L$:
\begin{equation}
	\argmin_z \expect{y}{L(z,\, y)} \label{eq:findmean}.
\end{equation}
For the $L_2$ loss $L(z,y) = (z-y)^2$, this minimum is found at the arithmetic mean of the observations:
\begin{equation}
z=\expect{y}{y} \label{eq:l2min}.
\end{equation} %
The $L_1$ loss, the sum of absolute deviations $L(z,y) = |z-y|$, in turn, has its optimum at the median of the observations.
The general class of deviation-minimizing estimators are known as M-estimators \cite{Huber1964}. From a statistical viewpoint, summary estimation using these common loss functions can be seen as ML estimation by interpreting the loss function as the negative log likelihood. %

Training neural network regressors is a generalization of this point estimation procedure. Observe the form of the typical training task for a set of input-target pairs $(x_i, y_i)$, where the network function $f_\theta(x)$ is parameterized by $\theta$:
\begin{equation}
	\argmin_\theta \expect{(x,y)}{L(f_\theta(x),\, y)} \label{eq:train_nn}.
\end{equation}
Indeed, if we remove the dependency on input data, and use a trivial $f_\theta$ that merely outputs a learned scalar, the task reduces to \eqref{eq:findmean}. Conversely, the full training task decomposes to the same minimization problem at every training sample; simple manipulations show that \eqref{eq:train_nn} is equivalent to
\begin{equation}
\argmin_\theta \expect{x}{\expect{y | x}{L(f_\theta(x),\, y)}}. \label{eq:train_nn_decomp}
\end{equation}
The network can, in theory, minimize this loss by solving the point estimation problem separately for each input sample. Hence, the properties of the underlying loss are inherited by neural network training.

The usual process of training regressors by Equation~\ref{eq:regression} over a finite number of input-target pairs $(x_i, y_i)$ hides a subtle point: instead of the 1:1 mapping between inputs and targets (falsely) implied by that process, in reality the mapping is multiple-valued. For example, in a superresolution task \cite{Ledig2016} over all natural images, a low-resolution image $x$ can be explained by many different high-resolution images $y$, as knowledge about the exact positions and orientations of the edges and texture is lost in decimation. In other words, $p(y|x)$ is the highly complex distribution of natural images consistent with the low-resolution $x$. Training a neural network regressor using training pairs of low- and high-resolution images using the $L_2$ loss, the network learns to output the average of all plausible explanations (e.g., edges shifted by different amounts), which results in spatial blurriness for the network's predictions. A significant amount of work has been done to combat this well known tendency, for example by using learned discriminator functions as losses \cite{Ledig2016,Isola2017}. %

Our observation is that for certain problems this tendency has an unexpected benefit.  A trivial, and, at first sight, useless, property of $L_2$ minimization is that on expectation, the estimate remains unchanged if we replace the targets with random numbers whose expectations match the targets. This is easy to see: Equation~\eqref{eq:l2min} holds, no matter what particular distribution the $y$s are drawn from. Consequently, the optimal network parameters $\theta$ of Equation~\eqref{eq:train_nn_decomp} also remain unchanged, if input-conditioned target distributions $p(y|x)$ are replaced with arbitrary distributions that have the same conditional expected values. \emph{This implies that we can, in principle, corrupt the training targets of a neural network with zero-mean noise without changing what the network learns.} Combining this with the corrupted inputs from Equation~\ref{eq:regression}, we are left with the empirical risk minimization task
\begin{equation}
\argmin_\theta \sum_i L\left( f_\theta(\inputx_i),\, \inputy_i \right), \label{eq:n2n}
\end{equation}
where both the inputs and the targets are now drawn from a corrupted distribution (not necessarily the same), conditioned on the underlying, unobserved clean target $y_i$ such that $\expectation{\inputy_i | \inputx_i} = y_i$. Given infinite data, the solution is the same as that of \eqref{eq:regression}. For finite data, the variance is the average variance of the corruptions in the targets, divided by the number of training samples (see \suppmat). %
Interestingly, none of the above relies on a likelihood model of the corruption, nor a density model (prior) for the underlying clean image manifold. That is, we do not need an explicit $p(\text{noisy}|\text{clean})$ or $p(\text{clean})$, as long as we have data distributed according to them.

In many image restoration tasks, the expectation of the corrupted input data \emph{is} the clean target that we seek to restore. Low-light photography is an example: a long, noise-free exposure is the average of short, independent, noisy exposures. With this in mind, the above suggests %
the ability to learn to remove photon noise given only pairs of noisy images, with no need for potentially expensive or difficult long exposures. 
Similar observations can be made about other loss functions. For instance, the $L_1$ loss recovers the median of the targets, %
meaning that neural networks can be trained to repair images with significant (up top 50\%) outlier content, again only requiring access to pairs of such corrupted images.

In the next sections, we present a wide variety of examples demonstrating that these theoretical capabilities are also efficiently realizable in practice. %

%% file: appendix-content.tex
\newcommand*{\movedown}[1]{\smash{\raisebox{-0.15ex}{#1}}}
\newcolumntype{L}{>{\collectcell\movedown}l<{\endcollectcell}}
\newcolumntype{R}{>{\collectcell\movedown}r<{\endcollectcell}}
\newcolumntype{C}{>{\collectcell\movedown}c<{\endcollectcell}}

\newcommand{\layer}[1]{\small\textsc{#1}}
\newcommand{\tblnetwork}{
\begin{table}[t]
\centering
\vspace*{\baselineskip}
\begin{tabular}{|L|C|L|}
\hline
\textsc{Name} & $N_\mathit{out}$ & \textsc{Function} \\
\hline
\raisebox{0mm}[3mm]{}%
\layer{input}       & $n$    & \\
\layer{enc\_conv0}  & $48$   & Convolution $3\times3$ \\
\layer{enc\_conv1}  & $48$   & Convolution $3\times3$ \\ 
\layer{pool1}       & $48$   & Maxpool $2\times2$ \\
\layer{enc\_conv2}  & $48$   & Convolution $3\times3$ \\ 
\layer{pool2}       & $48$   & Maxpool $2\times2$ \\
\layer{enc\_conv3}  & $48$   & Convolution $3\times3$ \\ 
\layer{pool3}       & $48$   & Maxpool $2\times2$ \\
\layer{enc\_conv4}  & $48$   & Convolution $3\times3$ \\ 
\layer{pool4}       & $48$   & Maxpool $2\times2$ \\
\layer{enc\_conv5}  & $48$   & Convolution $3\times3$ \\ 
\layer{pool5}       & $48$   & Maxpool $2\times2$ \\
\layer{enc\_conv6}  & $48$   & Convolution $3\times3$ \\ 
\layer{upsample5}   & $48$   & Upsample $2\times2$ \\
\layer{concat5}     & $96$   & Concatenate output of \layer{pool4} \\
\layer{dec\_conv5a} & $96$   & Convolution $3\times3$ \\ 
\layer{dec\_conv5b} & $96$   & Convolution $3\times3$ \\ 
\layer{upsample4}   & $96$   & Upsample $2\times2$ \\
\layer{concat4}     & $144$  & Concatenate output of \layer{pool3} \\
\layer{dec\_conv4a} & $96$   & Convolution $3\times3$ \\ 
\layer{dec\_conv4b} & $96$   & Convolution $3\times3$ \\ 
\layer{upsample3}   & $96$   & Upsample $2\times2$ \\
\layer{concat3}     & $144$  & Concatenate output of \layer{pool2} \\
\layer{dec\_conv3a} & $96$   & Convolution $3\times3$ \\ 
\layer{dec\_conv3b} & $96$   & Convolution $3\times3$ \\ 
\layer{upsample2}   & $96$   & Upsample $2\times2$ \\
\layer{concat2}     & $144$  & Concatenate output of \layer{pool1} \\
\layer{dec\_conv2a} & $96$   & Convolution $3\times3$ \\ 
\layer{dec\_conv2b} & $96$   & Convolution $3\times3$ \\ 
\layer{upsample1}   & $96$   & Upsample $2\times2$ \\
\layer{concat1}     & $96$$+$$n$ & Concatenate \layer{input} \\
\layer{dec\_conv1a} & $64$   & Convolution $3\times3$ \\ 
\layer{dec\_conv1b} & $32$   & Convolution $3\times3$ \\ 
\layer{dev\_conv1c} & $m$    & Convolution $3\times3$, linear act. \\
\hline
\end{tabular}
\caption{\label{tbl:network}%
Network architecture used in our experiments.
$N_\mathit{out}$ denotes the number of output feature maps for each layer.
Number of network input channels $n$ and output channels $m$ depend on the experiment. 
All convolutions use padding mode ``same'', and except for the last layer are followed by leaky ReLU activation function~\cite{Maas2013} with $\alpha=0.1$.
Other layers have linear activation.
Upsampling is nearest-neighbor.
}
\vfill
\end{table}
}

\appendixsection{Network architecture}
\tblnetwork

Table~\ref{tbl:network} shows the structure of the U-network~\cite{Ronneberger2015} used in all of our tests, with the exception of
the first test in Section~3.1 that used the ``RED30'' network~\cite{Mao2016b}.
For all basic noise and text removal experiments with RGB images, the number of input and output channels were \mbox{$n=m=3$}. For
Monte Carlo denoising we had \mbox{$n=9, m=3$}, i.e., input contained RGB pixel color, RGB albedo, and a 3D normal vector
per pixel. The MRI reconstruction was done with monochrome images (\mbox{$n=m=1$}).
Input images were represented in range $[-0.5, 0.5]$.

\appendixsection{Training parameters}

The network weights were initialized following He et al.~\yrcite{he2015}.
No batch normalization, dropout or other regularization techniques were used.
Training was done using ADAM~\cite{adam} with parameter values \mbox{$\beta_1=0.9$}, \mbox{$\beta_2=0.99$}, \mbox{$\epsilon=10^{-8}$}.

Learning rate was kept at a constant value during training except for a brief rampdown
period at where it was smoothly brought to zero.
Learning rate of $0.001$ was used for all experiments except Monte Carlo
denoising, where $0.0003$ was found to provide better stability. Minibatch size of~4 
was used in all experiments.

\appendixsection{Finite corrupted data in $L_2$ minimization}

\newcommand{\expectbig}[2]{\mathbb{E}_{#1}\left[{#2}\right]}
\newcommand{\eno}[2]{\mathbb{E}_{#1}{#2}}
\newcommand{\su}{( \sum_i y_i )}
\newcommand{\suh}{(\sum_i \hat y_i )}

Let us compute the expected error in $L_2$ norm minimization task when corrupted targets $\{\hat y_i\}_{i=1}^N$ are used in place of the clean targets $\{y_i\}_{i=1}^N$, with $N$ a finite number. Let $y_i$ be arbitrary random variables, such that $\expectation{\hat y_i} = y_i$. As usual, the point of least deviation is found at the respective mean. The expected squared difference between these means across realizations of the noise is then:
\begin{equation}
	\begin{split}
		& \expectbig{\hat y} {\frac{1}{N}\sum_i y_i - \frac{1}{N}\sum_i \hat y_i}^2 \\
		=& \frac{1}{N^2} \left[ \eno{\hat y}{\su^2} - 2 \expectbig{\hat y}{\su \suh} + \eno{\hat y}{\suh^2} \right] \\
		=& \frac{1}{N^2} \mathrm{Var} \suh \\
		=& \frac{1}{N} \left[\frac{1}{N} \sum_i \sum_j \mathrm{Cov}(\hat y_i, \hat y_j) \right]
	\end{split}
\end{equation}
In the intermediate steps, we have used $\eno{\hat y}{\suh} = \sum_i y_i$ and basic properties of (co)variance. If the corruptions are mutually uncorrelated, the last row simplifies to
\begin{equation}
	\begin{split}
		\frac{1}{N} \left[\frac{1}{N} \sum_i \mathrm{Var} (y_i) \right]
	\end{split}
\end{equation}
In either case, the variance of the estimate is the average (co)variance of the corruptions, divided by the number of samples $N$. Therefore, the error approaches zero as the number of samples grows. The estimate is unbiased in the sense that it is correct on expectation, even with a finite amount of data.

The above derivation assumes scalar target variables. When $\hat y_i$ are images, $N$ is to be taken as the total number of scalars in the images, i.e., \#images $\times$ \#pixels/image $\times$ \#color channels.

\appendixsection{Mode seeking and the ``$L_0$'' norm}

Interestingly, while the ``$L_0$ norm'' could intuitively be expected to converge to an exact mode, i.e. a local maximum of the probability density function of the data, theoretical analysis reveals that it recovers a slightly different point. While an actual mode is a zero-crossing of the \emph{derivative} of the PDF, the $L_0$ norm minimization recovers a zero-crossing of its \emph{Hilbert transform} instead. We have verified this behavior in a variety of numerical experiments, and, in practice, we find that the estimate is typically close to the true mode. This can be explained by the fact that the Hilbert transform approximates differentiation (with a sign flip): the latter is a multiplication by $i \omega$ in the Fourier domain, whereas the Hilbert transform is a multiplication by $-i ~ \mathrm{sgn}(\omega)$.

For a continuous data density $q(x)$, the norm minimization task for $L_p$ amounts to finding a point $x^*$ that has a minimal expected $p$-norm distance (suitably normalized, and omitting the $p$th root) from points $y \sim q(y)$:
\begin{equation}
	\begin{split}
		x^* &= \argmin_x \expect{y \sim q}{\frac{1}{p} |x-y|^p} \\
	&= \argmin_x \int \frac{1}{p} |x-y|^p q(y)\, \diff y
	\end{split}
\end{equation}
Following the typical procedure, the minimizer is found at a root of the derivative of the expression under argmin:
\begin{equation}
	\begin{split}
		0 &= \frac{\partial}{\partial{x}} \int \frac{1}{p} |x-y|^p q(y)\, \diff y \\
		&=  \int \mathrm{sgn}(x-y) |x-y|^{p-1} q(y)\, \diff y
	\end{split}
\end{equation}
This equality holds also when we take $\lim_{p\to 0}$.
The usual results for $L_2$ and $L_1$ norms can readily be derived from this form. For the $L_0$ case, we take $p=0$ and obtain
\begin{equation}
	\begin{split}
		0 &=  \int \mathrm{sgn}(x-y) |x-y|^{-1} q(y)\, \diff y \\
		&= \int \frac{1}{x-y} q(y)\, \diff y.
	\end{split}
\end{equation}
The right hand side is the formula for the Hilbert transform of $q(x)$, up to a constant multiplier.

%% file: paper.bbl
\begin{thebibliography}{27}
\providecommand{\natexlab}[1]{#1}
\providecommand{\url}[1]{\texttt{#1}}
\expandafter\ifx\csname urlstyle\endcsname\relax
  \providecommand{\doi}[1]{doi: #1}\else
  \providecommand{\doi}{doi: \begingroup \urlstyle{rm}\Url}\fi

\bibitem[Ashish~Bora(2018)]{ambientgan}
Ashish~Bora, Eric~Price, Alexandros G.~Dimakis.
\newblock {AmbientGAN: G}enerative models from lossy measurements.
\newblock \emph{ICLR}, 2018.

\bibitem[Cerd{\'{a}}{-}Company et~al.(2016)Cerd{\'{a}}{-}Company,
  P{\'{a}}rraga, and Otazu]{ToneMapSurvey}
Cerd{\'{a}}{-}Company, Xim, P{\'{a}}rraga, C.~Alejandro, and Otazu, Xavier.
\newblock Which tone-mapping operator is the best? {A} comparative study of
  perceptual quality.
\newblock \emph{CoRR}, abs/1601.04450, 2016.

\bibitem[Chaitanya et~al.(2017)Chaitanya, Kaplanyan, Schied, Salvi, Lefohn,
  Nowrouzezahrai, and Aila]{Chaitanya2017}
Chaitanya, Chakravarty R.~Alla, Kaplanyan, Anton~S., Schied, Christoph, Salvi,
  Marco, Lefohn, Aaron, Nowrouzezahrai, Derek, and Aila, Timo.
\newblock Interactive reconstruction of {Monte Carlo} image sequences using a
  recurrent denoising autoencoder.
\newblock \emph{ACM Trans. Graph.}, 36\penalty0 (4):\penalty0 98:1--98:12,
  2017.

\bibitem[Dabov et~al.(2007)Dabov, Foi, Katkovnik, and Egiazarian]{Dabov2007}
Dabov, K., Foi, A., Katkovnik, V., and Egiazarian, K.
\newblock Image denoising by sparse {3-D} transform-domain collaborative
  filtering.
\newblock \emph{IEEE Trans. Image Process.}, 16\penalty0 (8):\penalty0
  2080--2095, 2007.

\bibitem[Goodfellow et~al.(2014)Goodfellow, Pouget-Abadie, Mirza, Xu,
  Warde-Farley, Ozair, Courville, and Bengio]{Goodfellow2014}
Goodfellow, Ian, Pouget-Abadie, Jean, Mirza, Mehdi, Xu, Bing, Warde-Farley,
  David, Ozair, Sherjil, Courville, Aaron, and Bengio, Yoshua.
\newblock {Generative Adversarial Networks}.
\newblock In \emph{NIPS}, 2014.

\bibitem[Hasinoff et~al.(2016)Hasinoff, Sharlet, Geiss, Adams, Barron, Kainz,
  Chen, and Levoy]{Hasinoff2016}
Hasinoff, Sam, Sharlet, Dillon, Geiss, Ryan, Adams, Andrew, Barron,
  Jonathan~T., Kainz, Florian, Chen, Jiawen, and Levoy, Marc.
\newblock Burst photography for high dynamic range and low-light imaging on
  mobile cameras.
\newblock \emph{ACM Trans. Graph.}, 35\penalty0 (6):\penalty0 192:1--192:12,
  2016.

\bibitem[He et~al.(2015)He, Zhang, Ren, and Sun]{he2015}
He, Kaiming, Zhang, Xiangyu, Ren, Shaoqing, and Sun, Jian.
\newblock Delving deep into rectifiers: Surpassing human-level performance on
  imagenet classification.
\newblock \emph{CoRR}, abs/1502.01852, 2015.

\bibitem[Huber(1964)]{Huber1964}
Huber, Peter~J.
\newblock Robust estimation of a location parameter.
\newblock \emph{Ann. Math. Statist.}, 35\penalty0 (1):\penalty0 73--101, 1964.

\bibitem[Iizuka et~al.(2017)Iizuka, Simo-Serra, and Ishikawa]{Iizuka2017}
Iizuka, Satoshi, Simo-Serra, Edgar, and Ishikawa, Hiroshi.
\newblock Globally and locally consistent image completion.
\newblock \emph{ACM Trans. Graph.}, 36\penalty0 (4):\penalty0 107:1--107:14,
  2017.

\bibitem[Isola et~al.(2017)Isola, Zhu, Zhou, and Efros]{Isola2017}
Isola, Phillip, Zhu, Jun-Yan, Zhou, Tinghui, and Efros, Alexei~A.
\newblock Image-to-image translation with conditional adversarial networks.
\newblock In \emph{Proc. CVPR 2017}, 2017.

\bibitem[Kingma \& Ba(2015)Kingma and Ba]{adam}
Kingma, Diederik~P. and Ba, Jimmy.
\newblock Adam: {A} method for stochastic optimization.
\newblock In \emph{ICLR}, 2015.

\bibitem[Ledig et~al.(2017)Ledig, Theis, Huszar, Caballero, Aitken, Tejani,
  Totz, Wang, and Shi]{Ledig2016}
Ledig, Christian, Theis, Lucas, Huszar, Ferenc, Caballero, Jose, Aitken,
  Andrew~P., Tejani, Alykhan, Totz, Johannes, Wang, Zehan, and Shi, Wenzhe.
\newblock Photo-realistic single image super-resolution using a generative
  adversarial network.
\newblock In \emph{Proc. CVPR}, pp.\  105--114, 2017.

\bibitem[Lee et~al.(2017)Lee, Yoo, and Ye]{Lee2017}
Lee, D., Yoo, J., and Ye, J.~C.
\newblock Deep residual learning for compressed sensing {MRI}.
\newblock In \emph{Proc. IEEE 14th International Symposium on Biomedical
  Imaging (ISBI 2017)}, pp.\  15--18, 2017.

\bibitem[Lustig et~al.(2008)Lustig, Donoho, Santos, and Pauly]{Lustig2007}
Lustig, Michael, Donoho, David~L., Santos, Juan~M., and Pauly, John~M.
\newblock Compressed sensing {MRI}.
\newblock In \emph{{IEEE} Signal Processing Magazine}, volume~25, pp.\  72--82,
  2008.

\bibitem[Maas et~al.(2013)Maas, Hannun, and Ng]{Maas2013}
Maas, Andrew~L, Hannun, Awni~Y, and Ng, Andrew.
\newblock Rectifier nonlinearities improve neural network acoustic models.
\newblock In \emph{Proc. International Conference on Machine Learning (ICML)},
  volume~30, 2013.

\bibitem[Mao et~al.(2016)Mao, Shen, and Yang]{Mao2016b}
Mao, Xiao{-}Jiao, Shen, Chunhua, and Yang, Yu{-}Bin.
\newblock Image restoration using convolutional auto-encoders with symmetric
  skip connections.
\newblock In \emph{Proc. NIPS}, 2016.

\bibitem[Martin et~al.(2001)Martin, Fowlkes, Tal, and Malik]{BSD300}
Martin, D., Fowlkes, C., Tal, D., and Malik, J.
\newblock A database of human segmented natural images and its application to
  evaluating segmentation algorithms and measuring ecological statistics.
\newblock In \emph{Proc. ICCV}, volume~2, pp.\  416--423, 2001.

\bibitem[Mäkitalo \& Foi(2011)Mäkitalo and Foi]{Makitalo2011}
Mäkitalo, Markku and Foi, Alessandro.
\newblock Optimal inversion of the {Anscombe} transformation in low-count
  {Poisson} image denoising.
\newblock \emph{IEEE Trans. Image Process.}, 20\penalty0 (1):\penalty0 99--109,
  2011.

\bibitem[Reinhard et~al.(2002)Reinhard, Stark, Shirley, and
  Ferwerda]{Reinhard2002}
Reinhard, Erik, Stark, Michael, Shirley, Peter, and Ferwerda, James.
\newblock Photographic tone reproduction for digital images.
\newblock \emph{ACM Trans. Graph.}, 21\penalty0 (3):\penalty0 267--276, 2002.

\bibitem[Ronneberger et~al.(2015)Ronneberger, Fischer, and
  Brox]{Ronneberger2015}
Ronneberger, Olaf, Fischer, Philipp, and Brox, Thomas.
\newblock U-net: Convolutional networks for biomedical image segmentation.
\newblock \emph{MICCAI}, 9351:\penalty0 234--241, 2015.

\bibitem[Rousselle et~al.(2011)Rousselle, Knaus, and Zwicker]{Rousselle2011}
Rousselle, Fabrice, Knaus, Claude, and Zwicker, Matthias.
\newblock {Adaptive sampling and reconstruction using greedy error
  minimization}.
\newblock \emph{ACM Trans. Graph.}, 30\penalty0 (6):\penalty0 159:1--159:12,
  2011.

\bibitem[Srivastava et~al.(2014)Srivastava, Hinton, Krizhevsky, Sutskever, and
  Salakhutdinov]{srivastava2014}
Srivastava, Nitish, Hinton, Geoffrey, Krizhevsky, Alex, Sutskever, Ilya, and
  Salakhutdinov, Ruslan.
\newblock Dropout: A simple way to prevent neural networks from overfitting.
\newblock \emph{Journal of Machine Learning Research}, 15:\penalty0 1929--1958,
  2014.

\bibitem[Ulyanov et~al.(2017)Ulyanov, Vedaldi, and Lempitsky]{Ulyanov2017b}
Ulyanov, Dmitry, Vedaldi, Andrea, and Lempitsky, Victor~S.
\newblock Deep image prior.
\newblock \emph{CoRR}, abs/1711.10925, 2017.

\bibitem[Veach \& Guibas(1995)Veach and Guibas]{Veach1995}
Veach, Eric and Guibas, Leonidas~J.
\newblock Optimally combining sampling techniques for {M}onte {C}arlo
  rendering.
\newblock In \emph{Proc. ACM SIGGRAPH 95}, pp.\  419--428, 1995.

\bibitem[Wang et~al.(2016)Wang, Su, Ying, Peng, Zhu, Liang, Feng, and
  Liang]{Wang2016}
Wang, S., Su, Z., Ying, L., Peng, X., Zhu, S., Liang, F., Feng, D., and Liang,
  D.
\newblock Accelerating magnetic resonance imaging via deep learning.
\newblock In \emph{Proc. IEEE 13th International Symposium on Biomedical
  Imaging (ISBI)}, pp.\  514--517, 2016.

\bibitem[Zeyde et~al.(2010)Zeyde, Elad, and Protter]{Zeyde2010}
Zeyde, R., Elad, M., and Protter, M.
\newblock On single image scale-up using sparse-representations.
\newblock In \emph{Proc. Curves and Surfaces: 7th International Conference},
  pp.\  711--730, 2010.

\bibitem[Zhang et~al.(2016)Zhang, Isola, and Efros]{Zhang2016colorization}
Zhang, Richard, Isola, Phillip, and Efros, Alexei~A.
\newblock Colorful image colorization.
\newblock In \emph{Proc. ECCV}, pp.\  649--666, 2016.

\end{thebibliography}
